\documentclass[letterpaper]{article} 
\usepackage{nips12submit_e,times}
\usepackage{amsmath,amsthm,amssymb,dsfont,algorithm,algorithmic,url,subfig,multirow,array,graphics,graphicx,color,stmaryrd,hyperref}

\newtheorem{lem}{Lemma}
\newtheorem{thm}[lem]{Theorem}
\newtheorem{cor}[lem]{Corollary}

\newtheorem{defn}[lem]{Definition}

\newcommand{\R}{{\mathbb R}}
\newcommand{\RR}{{\cal R}}

\renewcommand{\H}{{\cal H}}
\newcommand{\X}{{\cal X}}
\newcommand{\D}{{\cal D}}
\renewcommand{\L}{{\cal L}}
\newcommand{\T}{{\cal T}}
\newcommand{\Y}{{\cal Y}}
\newcommand{\A}{{\cal A}}
\newcommand{\F}{{\cal F}}

\newcommand{\Q}{{\cal Q}}
\renewcommand{\P}{{\cal P}}

\newcommand{\bc}[1]{\left\{{#1}\right\}}
\newcommand{\br}[1]{\left({#1}\right)}
\newcommand{\bs}[1]{\left[{#1}\right]}
\newcommand{\abs}[1]{\left| {#1} \right|}
\newcommand{\norm}[1]{\left\| {#1} \right\|}

\newcommand{\bsd}[1]{\left\llbracket{#1}\right\rrbracket}

\renewcommand{\O}[1]{{\cal O}\br{{#1}}}
\newcommand{\softO}[1]{\tilde{\cal O}\br{{#1}}}
\newcommand{\Om}[1]{\Omega\br{{#1}}}
\newcommand{\E}[1]{{\mathbb E}\bsd{{#1}}}
\newcommand{\EE}[2]{\underset{#1}{\mathbb E}\bsd{{#2}}}

\renewcommand{\Pr}[1]{{\mathbb P}\bs{{#1}}}
\newcommand{\Prr}[2]{\underset{#1}{\mathbb P}\bs{{#2}}}
\newcommand{\ip}[2]{\left\langle{#1},{#2}\right\rangle}

\renewcommand{\vec}[1]{{\mathbf{#1}}}
\newcommand{\vecx}{\vec{x}}
\newcommand{\x}{\vec{x}}
\newcommand{\vecy}{\vec{y}}
\newcommand{\vecz}{\vec{z}}
\newcommand{\vecxy}{\vecx,\vecy}
\newcommand{\vecw}{\vec{w}}
\newcommand{\vecW}{\vec{W}}

\newcommand{\vecr}{\vec{r}}
\newcommand{\vecs}{\vec{s}}

\newcommand{\veczero}{\vec{0}}
\newcommand{\vectheta}{\text{\boldmath$\theta$}}

\newcommand{\eil}[2]{\ell_{#1}\br{#2}}
\newcommand{\eild}[1]{\eil{\epsilon}{#1}}
\newcommand{\ellac}[2]{\eil{\text{actual}}{#1,#2}}
\newcommand{\gml}[2]{\bs{#2}_{#1}}
\newcommand{\gmld}[1]{\gml{\gamma}{#1}}
\newcommand{\pos}[1]{\gml{+}{#1}}
\newcommand{\evbad}[1]{\textsf{BAD-APPROX}\br{#1}}

\newcommand{\lord}[1]{\ell_{\text{ord}}\br{#1}}
\newcommand{\lndcg}[1]{\ell_{\text{NDCG}}\br{#1}}
\newcommand{\pmone}{\bc{-1,+1}}
\newcommand{\ind}{\mathds{1}}
\newcommand{\lsq}[1]{\ell_{\text{sq}}\br{#1}}
\newcommand{\sign}{\text{sign}}

\title{Supervised Learning with Similarity Functions}

\author{
Purushottam Kar\\
Indian Institute of Technology\\
Kanpur, INDIA\\
\texttt{purushot@cse.iitk.ac.in}
\And
Prateek Jain\\
Microsoft Research Lab\\
Bangalore, INDIA\\
\texttt{prajain@microsoft.com}
}

\date{}

\nipsfinalcopy 

\begin{document}

\maketitle

\begin{abstract}
We address the problem of general supervised learning when data can only be accessed through an (indefinite) similarity function between data points.
Existing work on learning with indefinite kernels has concentrated solely on binary/multi-class classification problems. We propose a model that is generic enough to handle any supervised learning task and also subsumes the model previously proposed for classification. We give a ``goodness'' criterion for similarity functions w.r.t. a given supervised learning task and then adapt a well-known landmarking technique to provide efficient algorithms for supervised learning using ``good'' similarity functions. We demonstrate the effectiveness of our model on three important supervised learning problems: a) real-valued regression, b) ordinal regression and c) ranking where we show that our method guarantees bounded generalization error. Furthermore, for the case of real-valued regression, we give a natural goodness definition that, when used in conjunction with a recent result in sparse vector recovery, guarantees a sparse predictor with bounded generalization error. 
Finally, we report results of our learning algorithms on regression and ordinal regression tasks using non-PSD similarity functions and demonstrate the effectiveness of our algorithms, especially that of the sparse landmark selection algorithm that achieves significantly higher accuracies than the baseline methods while offering reduced computational costs.
\end{abstract}
\section{Introduction}
The goal of this paper is to develop an extended framework for supervised learning with similarity functions. Kernel learning algorithms \cite{kernel-book} have become the mainstay of discriminative learning with an incredible amount of effort having been put in, both from the theoretician's as well as the practitioner's side. 
However, these algorithms typically require the similarity function to be a positive semi-definite (PSD) function, which can be a limiting factor for several applications. Reasons being: 1) the Mercer's condition is a formal statement that is hard to verify, 2) several natural notions of similarity that arise in practical scenarios are not PSD, and 3) it is not clear as to why an artificial constraint like PSD-ness should limit the usability of a kernel. 

Several recent papers have demonstrated that indefinite similarity functions can indeed be successfully used for learning \cite{feature-svm-indef,learn-indef,learn-kern-from-indef,svm-indef}. However, most of the existing work focuses on classification tasks and provides specialized techniques for the same, albeit with little or no theoretical guarantees. A notable exception is the line of work by \cite{learn-sim,learn-dissim,dissim-nips} that defines a goodness criterion for a similarity function and then provides an algorithm that can exploit this goodness criterion to obtain provably accurate classifiers. However, their definitions are yet again restricted to the problem of classification as they take a ``margin'' based view of the problem that requires positive points to be more similar to positive points than to negative points by at least a constant margin. 

In this work, we instead take a ``target-value'' point of view and require that target values of similar points be similar. Using this view, we propose a generic goodness definition that also admits the goodness definition of \cite{learn-sim} for classification as a special case. Furthermore, our definition can be seen as imposing the existence of a smooth function over a generic space defined by similarity functions, rather than over a Hilbert space as required by typical goodness definitions of PSD kernels. 

We then adapt the landmarking technique of \cite{learn-sim} to provide an efficient algorithm that reduces learning tasks to corresponding learning problems over a linear space. The main technical challenge at this stage is to show that such reductions are able to provide good generalization error bounds for the learning tasks at hand. To this end, we consider three specific problems: a) regression, b) ordinal regression, and c) ranking. For each problem, we define appropriate surrogate loss functions, and show that our algorithm is able to, for each specific learning task, guarantee bounded generalization error with polynomial sample complexity. Moreover, by adapting a general framework given by \cite{kernel-sim-compare}, we show that these guarantees do not require the goodness definition to be overly restrictive by showing that our definitions admit all good PSD kernels as well.


For the problem of real-valued regression, we additionally provide a goodness definition that captures the intuition that usually, only a small number of landmarks are influential w.r.t. the learning task. However, to recover these landmarks, the uniform sampling technique would require sampling a large number of landmarks thus increasing the training/test time of the predictor. We address this issue by applying a sparse vector recovery algorithm given by \cite{acc-sparse} and show that the resulting sparse predictor still has bounded generalization error.   

We also address an important issue faced by algorithms that use landmarking as a feature constructions step viz \cite{learn-sim,learn-dissim,dissim-nips}, namely that they typically assume separate landmark and training sets for ease of analysis. In practice however, one usually tries to overcome paucity of training data by reusing training data as landmark points as well. We use an argument outlined in \cite{double-dipping} to theoretically justify such
``double dipping'' in our case. The details of the argument are given in \ref{app:double-dipping}.

We perform several experiments on benchmark datasets that demonstrate significant performance gains for our methods over the baseline of kernel regression. Our sparse landmark selection technique provides significantly better predictors that are also more efficient at test time. 


{\bf Related Work}: Existing approaches to extend kernel learning algorithms to indefinite kernels can be classified into three broad categories: a) those that use indefinite kernels directly with existing kernel learning algorithms, resulting in non-convex formulations \cite{feature-svm-indef,learn-indef}. 
 b) those that convert a given indefinite kernel into a PSD one by either projecting onto the PSD-cone \cite{learn-kern-from-indef,svm-indef} or performing other spectral operations \cite{classfn-sim}. The second approach is usually expensive due to the spectral operations involved apart from making the method inherently transductive. 
 Moreover, any domain knowledge stored in the original kernel is lost due to these task oblivious operations and consequently, no generalization guarantees can be given.
 c) those that use notions of ``task-kernel alignment'' or equivalently, notions of ``goodness'' of a kernel, to give learning algorithms 
 \cite{learn-sim,learn-dissim,dissim-nips}. This approach enjoys several advantages over the other approaches listed above. These models are able to use the indefinite kernel directly with existing PSD kernel learning techniques; all the while retaining the ability to give generalization bounds that 
 quantitatively parallel those of PSD kernel learning models. In this paper, we adopt the third approach for general supervised learning problem. 

\section{Problem formulation and Preliminaries}
\label{sec:formulation}
The goal in similarity-based supervised learning is to closely approximate a \emph{target} predictor $y : \X \rightarrow \Y$ over some domain $\X$ using a \emph{hypothesis} $\hat f(\ \cdot\ ;K) : \X \rightarrow \Y$ that restricts its interaction with data points to computing similarity values given by $K$.
Now, if the similarity function $K$ is not discriminative enough for the given task then we cannot hope to construct a predictor out of it that enjoys good generalization properties. Hence, it is natural to define the ``goodness'' of a given similarity function with respect to the learning task at hand. 
\begin{defn}[Good similarity function: preliminary]
\label{strong-good}
Given a learning task $y : \X \rightarrow \Y$ over some distribution $\D$, a similarity function $K : \X \times \X \rightarrow \R$ is said to be $\br{\epsilon_0,B}$-good with respect to this task if there exists some bounded weighing function $w : \X \rightarrow \bs{-B,B}$ such that for at least a $\br{1 - \epsilon_0}$ $\D$-fraction of the domain, we have $\ y(\vecx) = \EE{\vecx'\sim\D}{w(\vecx')y(\vecx')K(\vecx,\vecx')}.$
\end{defn}
The above definition is inspired by the definition of a ``good'' similarity function with respect to classification tasks given in \cite{learn-sim}. However, their definition is tied to class labels and thus applies only to classification tasks. Similar to \cite{learn-sim}, the above definition calls a similarity function $K$ ``good'' if the target value $y(\x)$ of a given point $\x$ can be approximated in terms of (a weighted combination of) the target values of the  $K$-``neighbors'' of $\x$. Also, note that this definition automatically enforces a smoothness prior on the framework.

However the above definition is too rigid. Moreover, it defines goodness in terms of violations, a non-convex loss function. To remedy this, we propose an alternative definition that incorporates an arbitrary (but in practice always convex) loss function. 
\begin{defn}[Good similarity function: final]
\label{weak-good}
Given a learning task $y : \X \rightarrow \Y$ over some distribution $\D$, a similarity function $K$ is said to be $\br{\epsilon_0,B}$-good with respect to a loss function $\ell_S : \R \times \Y \rightarrow \R$ if there exists some bounded weighing function $w : \X \rightarrow \bs{-B,B}$ such that if we define a predictor as $f(\x) := \EE{\vecx'\sim\D}{w(\vecx')K(\vecx,\vecx')}$, then we have $\EE{\vecx \sim \D}{\ell_S(f(\vecx),y(\vecx))} \leq \epsilon_0$.
\end{defn}
Note that Definition~\ref{weak-good} reduces to Definition~\ref{strong-good} for $\ell_S(a,b) = \ind_{\bc{a \neq b}}$. Moreover, for the case of binary classification where $y \in \pmone$, if we take $\ell_S(a,b) = \ind_{\bc{ab \leq B\gamma}}$, then we recover the $\br{\epsilon_0,\gamma}$-goodness definition of a similarity function, given in Definition 3 of \cite{learn-sim}. 
 Also note that, assuming $\underset{\vecx \in \X}{\sup}\bc{\abs{y(\vecx)}} < \infty$ we can w.l.o.g. merge $w(\vecx')y(\vecx')$ into a single term $w(\x')$. 

Having given this definition we must make sure that ``good'' similarity functions allow the construction of effective predictors (Utility property). Moreover, we must make sure that the definition does not exclude commonly used PSD kernels (Admissibility property). 
Below, we formally define these two properties and in later sections, show that for each of the learning tasks considered, our goodness definition satisfies these two properties. 
\subsection{Utility}
\label{sec:utility}
\begin{defn}[Utility]
A similarity function $K$ is said to be $\epsilon_0$-useful w.r.t. a loss function $\ellac{\cdot}{\cdot}$ if the following holds: there exists a learning algorithm $\A$ that, for any $\epsilon_1, \delta > 0$, when given $\text{poly}(1/\epsilon_1, \log(1/\delta))$ ``labeled'' and ``unlabeled'' samples from the input distribution $\D$, with probability at least $1-\delta$ , generates a hypothesis $\hat{f}(\x;K)$ s.t. $\EE{\vecx \sim \D}{\ellac{\hat f(\vecx)}{y(\vecx)}} \leq \epsilon_0 + \epsilon_1$. Note that $\hat{f}(\x;K)$ is restricted to access the data solely through $K$. 
\end{defn}
Here, the $\epsilon_0$ term captures the \emph{misfit} or the bias of the similarity function with respect to the learning problem. Notice that the above utility definition allows for learning from unlabeled data points and thus puts our approach in the semi-supervised learning framework. 

All our utility guarantees proceed by first using unlabeled samples as \emph{landmarks} to construct a landmarked space. Next, using the goodness definition, we show the existence of a good linear predictor in the landmarked space. This guarantee is obtained in two steps as outlined in Algorithm~\ref{alg-learn-sim-gen}: first of all we choose $d$ unlabeled landmark points and construct a map $\Psi : \X \rightarrow \R^d$ (see Step~\ref{stp:land} of Algorithm~\ref{alg-learn-sim-gen}) and show that 
 there exists a linear predictor over $\R^d$ that closely approximates the predictor $f$ used in Definition~\ref{weak-good} (see Lemma~\ref{landmark-guarantee} in \ref{app:general}). In the second step, we learn a predictor (over the landmarked space) using ERM over a fresh labeled training set (see Step~\ref{stp:erm} of Algorithm~\ref{alg-learn-sim-gen}). We then use individual task-specific arguments and Rademacher average-based generalization bounds \cite{rademacher-averages} thus proving the utility of the similarity function.
		\begin{algorithm}[t]
			\caption{\small Supervised learning with Similarity functions}
			\label{alg-learn-sim-gen}
			\begin{algorithmic}[1]
				\small{
					\REQUIRE A target predictor $y : \X \rightarrow \Y$ over a distribution $\D$, an $\br{\epsilon_0,B}$-good similarity function $K$, labeled training points sampled from $\D$: $\T = \bc{(\vecx^t_1,y_1),\ldots,(\vecx^t_n,y_n)}$, loss function $\ell_S : \R \times \Y \rightarrow \R^+$.
					\ENSURE A predictor $\hat f : \X \rightarrow \R$ with bounded true loss over $\D$
					\STATE Sample $d$ unlabeled landmarks from $\D$: $\L = \bc{\vecx^l_1,\ldots,\vecx^l_d}$ \label{stp:land}
         	\newline\texttt{// Else subsample $d$ landmarks from $\T$ (see \ref{app:double-dipping} for details)}
					\STATE $\Psi_\L : \vecx \mapsto 1/\sqrt{d}\br{K(\vecx,\vecx^l_1),\ldots,K(\vecx,\vecx^l_d)} \in \R^d$
					\STATE $\hat\vecw = \underset{\vecw \in \R^d : \norm{\vecw}_2 \leq B}{\arg\min}\sum_i^n\ell_S\br{\ip{\vecw}{\Psi_\L(\vecx^t_i)},y_i}$\label{stp:erm}
					\RETURN{$\hat f : \vecx \mapsto \ip{\hat\vecw}{\Psi_\L(\vecx)}$}
				}
			\end{algorithmic}
		\end{algorithm}
\subsection{Admissibility}
\label{sec:admissibility}
In order to show that our models are not too rigid, we would prove that they admit good PSD kernels. 
The notion of a good PSD kernel for us will be one that corresponds to a prevalent large margin technique for the given problem. In general, most notions correspond to the existence of a linear operator in the RKHS of the kernel that has small loss at large margin. More formally,
\begin{defn}[Good PSD Kernel]
\label{kernel-good}
Given a learning task $y : \X \rightarrow \Y$ over some distribution $\D$, a PSD kernel $K : \X \times \X \rightarrow \R$ with associated RKHS $\H_K$ and canonical feature map $\Phi_K : \X \rightarrow \H_K$ is said to be $\br{\epsilon_0,\gamma}$-good with respect to a loss function $\ell_K : \R \times \Y \rightarrow \R$ if there exists $\vecW^\ast \in \H_K$ such that $\norm{\vecW^\ast} = 1$ and
\[
\EE{\vecx \sim \D}{\ell_K\br{\frac{\ip{\vecW^\ast}{\Phi_K(\vecx)}}{\gamma},y(\vecx)}} < \epsilon_0
\]
\end{defn}
We will show, for all the learning tasks considered, that every $\br{\epsilon_0,\gamma}$-good PSD kernel, when treated as simply a similarity function with no consideration of its RKHS, is also $\br{\epsilon + \epsilon_1,B}$-good for arbitrarily small $\epsilon_1$  with $B = h(\gamma,\epsilon_1)$ for some function $h$.
To prove these results we will adapt techniques introduced in \cite{kernel-sim-compare} with certain modifications and task-dependent arguments. 
\section{Applications}
We will now instantiate the general learning model described above to real-valued regression, ordinal regression and ranking by providing utility and admissibility guarantees. Due to lack of space, we relegate all proofs as well as the discussion on ranking to the supplementary material (\ref{app:rank}).
\subsection{Real-valued Regression}
Real-valued regression is a quintessential learning problem \cite{kernel-book} that has received a lot of attention in the learning literature. In the following we shall present algorithms for performing real-valued regression using non-PSD similarity measures. We consider the problem with $\ellac{a}{b} = \abs{a-b}$ as the true loss function. For the surrogates $\ell_S$ and $\ell_K$, we choose the $\epsilon$-insensitive loss function \cite{kernel-book} defined as follows:
\[
\eild{a,b} = \eild{a-b} = \left\{
	\begin{array}{l l}
		0, & \quad \text{if $\abs{a-b} < \epsilon$},\\
    	\abs{a-b} - \epsilon, & \quad \text{otherwise}.\\
	\end{array}\right.
\]
The above loss function automatically gives us notions of good kernels and similarity functions by appealing to Definitions~\ref{kernel-good} and \ref{weak-good} respectively. It is easy to transfer error bounds in terms of absolute error to those in terms of mean squared error (MSE), a commonly used performance measure for real-valued regression. See \ref{app:sp_reg} for further discussion on the choice of the loss function.

Using the landmarking strategy described in Section~\ref{sec:utility}, we can reduce the problem of real regression to that of a linear regression problem in the landmarked space. More specifically, the ERM step in Algorithm~\ref{alg-learn-sim-gen} becomes the following: $\underset{\vecw \in \R^d : \norm{\vecw}_2 \leq B}{\arg\min}\sum_i^n\eild{\ip{\vecw}{\Psi_\L(\vecx_i)}-y_i}$.

There exist solvers (for instance \cite{svr-lin}) to efficiently solve the above problem on linear spaces. Using proof techniques sketched in Section~\ref{sec:utility} along with specific arguments for the $\epsilon$-insensitive loss, we can prove generalization guarantees and hence utility guarantees for the similarity function.

\begin{thm}
\label{good-sim-guarantee}
Every similarity function that is $\br{\epsilon_0,B}$-good for a regression problem with respect to the insensitive loss function $\eild{\cdot,\cdot}$ is $\br{\epsilon_0 + \epsilon}$-useful with respect to absolute loss as well as $\br{B\epsilon_0 + B\epsilon}$-useful with respect to mean squared error. Moreover, both the dimensionality of the landmarked space as well as the labeled sample complexity can be bounded by $\O{\frac{B^2}{\epsilon_1^2}\log{\frac{1}{\delta}}}$.
\end{thm}

%

We are also able to prove the following (tight) admissibility result:

\begin{thm}
\label{reg-psd-sim-good}
Every PSD kernel that is $\br{\epsilon_0,\gamma}$-good for a regression problem is, for any $\epsilon_1 > 0$, $\br{\epsilon_0 + \epsilon_1, \O{\frac{1}{\epsilon_1 \gamma^2}}}$-good as a similarity function as well. Moreover, for any $\epsilon_1 < 1/2$ and any $\gamma < 1$, there exists a regression instance and a corresponding kernel that is $\br{0,\gamma}$-good for the regression problem but only $\br{\epsilon_1,B}$-good as a similarity function for $B = \Om{\frac{1}{\epsilon_1\gamma^2}}$.
\end{thm}

\subsection{Sparse regression models}
\label{sec-sparse-reg}
An artifact of a random choice of landmarks is that very few of them might turn out to be ``informative'' with respect to the prediction problem at hand. For instance, in a network, there might exist \emph{hubs} or \emph{authoritative} nodes that yield rich information about the learning problem. If the relative abundance of such nodes is low then random selection would compel us to choose a large number of landmarks before enough ``informative'' ones have been collected.

However this greatly increases training and testing times due to the increased costs of constructing the landmarked space. Thus, the ability to prune away irrelevant landmarks would speed up training and test routines. We note that this issue has been addressed before in literature \cite{dissim-nips,classfn-sim} by way of landmark selection heuristics. In contrast, we guarantee that our predictor will select a small number of landmarks while incurring bounded generalization error. 
However this requires a careful restructuring of the learning model to incorporate the ``informativeness'' of landmarks. 

\begin{defn}
\label{reg-good-sim-sparse}
A similarity function $K$ is said to be $\br{\epsilon_0,B,\tau}$-good for a real-valued regression problem $y : \X \rightarrow \R$ if for some bounded weight function $w : \X \rightarrow \bs{-B,B}$ and \emph{choice function} $R : \X \rightarrow \bc{0,1}$  with $\EE{\vecx \sim \D}{R(\vecx)} = \tau$,
the predictor $f : \vecx \mapsto \EE{\vecx' \sim \D}{w(\vecx')K(\vecx,\vecx') | R(\vecx')}$ 
has bounded $\epsilon$-insensitive loss i.e. $\EE{\vecx \sim \D}{\eild{f(\vecx),y(\vecx)}} < \epsilon_0$.
\end{defn}

The role of the choice function is to single out informative landmarks, while $\tau$ specifies the relative density of informative landmarks.
Note that the above definition is similar in spirit to the goodness definition presented in \cite{learn-sim-improv}. While the motivation behind \cite{learn-sim-improv} was to give an improved admissibility result for binary classification, we squarely focus on the utility guarantees; with the aim of accelerating our learning algorithms via landmark pruning. 

We prove the utility guarantee in three steps as outlined in \ref{app:sp_reg}. First, we use the usual landmarking step to project the problem onto a linear space. This step guarantees the following:

\begin{thm}
\label{sparse-landmark-guarantee}
Given a similarity function that is $\br{\epsilon_0,B,\tau}$-good for a regression problem, there exists a randomized map $\Psi : \X \rightarrow \R^d$ for $d = \O{\frac{B^2}{\tau\epsilon_1^2}\log\frac{1}{\delta}}$ such that with probability at least $1-\delta$, there exists a linear operator $\tilde f : \vecx \mapsto \ip{\vecw}{\vecx}$ over $\R^d$ such that $\norm{\vecw}_1 \leq B$ with $\epsilon$-insensitive loss bounded by $\epsilon_0 + \epsilon_1$. Moreover, with the same confidence we have $\norm{\vecw}_0 \leq \frac{3d\tau}{2}$.
\end{thm}
Our proof follows that of \cite{learn-sim-improv}, however we additionally prove sparsity of $\vecw$ as well. The number of landmarks required here is a $\Om{1/\tau}$ fraction greater than that required by Theorem~\ref{good-sim-guarantee}. This formally captures the intuition presented earlier of a small fraction of dimensions (read landmarks) being actually relevant to the learning problem. So, in the second step, we use the \emph{Forward Greedy Selection} algorithm given in \cite{acc-sparse} to learn a sparse predictor. 
The use of this learning algorithm necessitates the use of a different generalization bound in the final step to complete the utility guarantee given below. 
We refer the reader to \ref{app:sp_reg} for the details of the algorithm and its utility analysis. 

\begin{thm}
\label{good-sim-sparse-guarantee}
Every similarity function that is $\br{\epsilon_0,B,\tau}$-good for a regression problem with respect to the insensitive loss function $\eild{\cdot,\cdot}$ is $\br{\epsilon_0 + \epsilon}$-useful with respect to absolute loss as well; with the dimensionality of the landmarked space being bounded by $\O{\frac{B^2}{\tau\epsilon_1^2}\log{\frac{1}{\delta}}}$ and the labeled sampled complexity being bounded by $\O{\frac{B^2}{\epsilon_1^2}\log{\frac{B}{\epsilon_1\delta}}}$. Moreover, this utility can be achieved by an $\O{\tau}$-sparse predictor on the landmarked space.
\end{thm}

We note that the improvements obtained here by using the sparse learning methods of \cite{acc-sparse} provide $\Om{\tau}$ increase in sparsity.
We now prove admissibility results for this sparse learning model. We do this by showing that the dense model analyzed in Theorem~\ref{good-sim-guarantee} and that given in Definition~\ref{reg-good-sim-sparse} are interpretable in each other for an appropriate selection of parameters. The guarantees in Theorem~\ref{reg-psd-sim-good} can then be invoked to conclude the admissibility proof.

\begin{thm}
Every $\br{\epsilon_0,B}$-good similarity function $K$ is also $\br{\epsilon_0,B,\frac{\bar w}{B}}$-good where $\bar w = \EE{\vecx \sim \D}{\abs{w(\vecx)}}$. Moreover, every $\br{\epsilon_0,B,\tau}$-good similarity function $K$ is also $\br{\epsilon_0,B/\tau}$-good.
\label{thm:util_sparse_reg}
\end{thm}

Using Theorem~\ref{reg-psd-sim-good}, we immediately have the following corollary:
\begin{cor}
Every PSD kernel that is $\br{\epsilon_0,\gamma}$-good for a regression problem is, for any $\epsilon_1 > 0$, $\br{\epsilon_0 + \epsilon_1, \O{\frac{1}{\epsilon_1 \gamma^2}},1}$-good as a similarity function as well.
\end{cor}

\subsection{Ordinal Regression}
\label{sec:oreg}
The problem of ordinal regression requires an accurate prediction of (discrete) labels coming from a finite ordered set $\bs{r} = \bc{1,2,\ldots,r}$. The problem is similar to both classification and regression, but has some distinct features due to which it has received independent attention \cite{ordreg-svm, ordreg-guarantee} in domains such as product ratings etc. 
The most popular performance measure for this problem is the absolute loss which is the absolute difference between the predicted and the true labels.

A natural and rather tempting way to solve this problem is to relax the problem to real-valued regression and threshold the output of the learned real-valued predictor using predefined thresholds $b_1,\ldots,b_r$ to get discrete labels. Although this approach has been prevalent in literature \cite{ordreg-guarantee}, as the discussion in the supplementary material shows, this leads to poor generalization guarantees in our model. More specifically, a goodness definition constructed around such a direct reduction is only able to ensure $\br{\epsilon_0 + 1}$-utility i.e. the absolute error rate is always greater than $1$.

One of the reasons for this is the presence of the thresholding operation that makes it impossible to distinguish between instances that would not be affected by small perturbations to the underlying real-valued predictor and those that would. To remedy this, we enforce a (soft) margin with respect to thresholding that makes the formulation more robust to noise. More formally, we expect that if a point belongs to the label $i$, then in addition to being sandwiched between the thresholds $b_i$ and $b_{i+1}$, it should be separated from these by a margin as well i.e. $b_i + \gamma \leq f(\vecx) \leq b_{i+1} - \gamma$.

This is a direct generalization of the margin principle in classification where we expect $\vecw^\top\vecx > b + \gamma$ for positively labeled points and $\vecw^\top\vecx < b - \gamma$ for negatively labeled points. Of course, wherein classification requires a single threshold, we require several, depending upon the number of labels. For any $x \in \R$, let $\pos{x} = \max\bc{x,0}$. Thus, if we define the $\gamma$-margin loss function to be $\gmld{x} := \pos{\gamma - x}$ (note that this is simply the well known hinge loss function scaled by a factor of $\gamma$), we can define our goodness criterion as follows:
\begin{defn}
\label{def:oreg-good}
A similarity function $K$ is said to be $\br{\epsilon_0,B}$-good for an ordinal regression problem $y : \X \rightarrow \bs{r}$ if for some bounded weight function $w : \X \rightarrow \bs{-B,B}$ and some (unknown but fixed) set of thresholds $\bc{b_i}_{i=1}^{r}$ with $b_1 = -\infty$, the predictor $f : \vecx \mapsto \EE{\vecx' \sim \D}{w(\vecx')K(\vecx,\vecx')}$ satisfies $\EE{\vecx \sim \D}{\gmld{f(\vecx) - b_{y(\vecx)}} + \gmld{b_{y(\vecx) + 1} - f(\vecx)}} < \epsilon_0$.
\end{defn}
%
%
We now give utility guarantees for our learning model. We shall give guarantees on both the misclassification error as well as the absolute error of our learned predictor. We say that a set of points $x_1, \ldots, x_i \ldots$ is $\Delta$-spaced if $\underset{i \neq j}{\min}\bc{\abs{x_i - x_j}} \geq \Delta$. Define the function $\psi_\Delta(x) = \frac{x + \Delta - 1}{\Delta}$.
\begin{thm}
\label{good-sim-guarantee-ord}
Let $K$ be a similarity function that is $\br{\epsilon_0,B}$-good for an ordinal regression problem with respect to $\Delta$-spaced thresholds and $\gamma$-margin loss. Let $\bar\gamma = \max\bc{\gamma,1}$. Then $K$ is $\psi_{\br{\Delta/\bar\gamma}}\br{\frac{\epsilon_0}{\bar\gamma}}$-useful with respect to ordinal regression error (absolute loss). Moreover, $K$ is $\br{\frac{\epsilon_0}{\bar\gamma}}$-useful with respect to the zero-one mislabeling error as well.
\end{thm}
We can bound, both dimensionality of the landmarked space as well as labeled sampled complexity, by $\O{\frac{B^2}{\epsilon_1^2}\log{\frac{1}{\delta}}}$. Notice that for $\epsilon_0 < 1$ and large enough $d,n$, we can ensure that the ordinal regression error rate is also bounded above by $1$ since $\underset{x \in \bs{0,1},\Delta > 0}{\sup}\br{\psi_\Delta\br{x}} = 1$. This is in contrast with the direct reduction to real valued regression which has ordinal regression error rate bounded \emph{below} by $1$. 
This indicates the advantage of the present model over a naive reduction to regression.
%

We can show that our definition of a good similarity function admits all good PSD kernels as well. The kernel goodness criterion we adopt corresponds to the large margin framework proposed by \cite{ordreg-svm}. We refer the reader to Appendix~\ref{app:oreg_adm} for the definition and give the admissibility result below.
\begin{thm}
\label{thm:adm_oreg}
Every PSD kernel that is $\br{\epsilon_0,\gamma}$-good for an ordinal regression problem is also $\br{\gamma_1\epsilon_0 + \epsilon_1, \O{\frac{\gamma_1^2}{\epsilon_1 \gamma^2}}}$-good as a similarity function with respect to the $\gamma_1$-margin loss for any $\gamma_1, \epsilon_1 > 0$. Moreover, for any $\epsilon_1 < \gamma_1/2$, there exists an ordinal regression instance and a corresponding kernel that is $\br{0,\gamma}$-good for the ordinal regression problem but only $\br{\epsilon_1,B}$-good as a similarity function with respect to the $\gamma_1$-margin loss function for $B = \Om{\frac{\gamma_1^2}{\epsilon_1\gamma^2}}$.
\end{thm}


Due to lack of space we refer the reader to \ref{app:rank} for a discussion on ranking models that includes utility and admissibility guarantees with respect to the popular NDCG loss.
\section{Experimental Results}
\label{sec:exps}
\begin{figure*}[t]
	\vspace*{-2ex}
	\centering
	\subfloat[Mean squared error for landmarking (RegLand), sparse landmarking (RegLand-Sp) and kernel regression (KR)]{
		\label{fig:reg}
		\includegraphics[width=0.35\textwidth, angle=0]{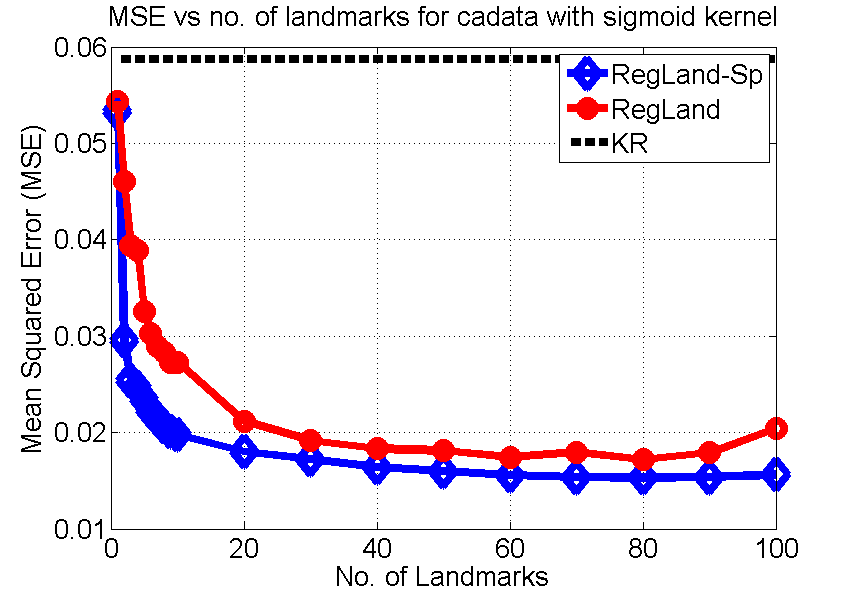}
		\hspace*{-2ex}
		\includegraphics[width=0.35\textwidth, angle=0]{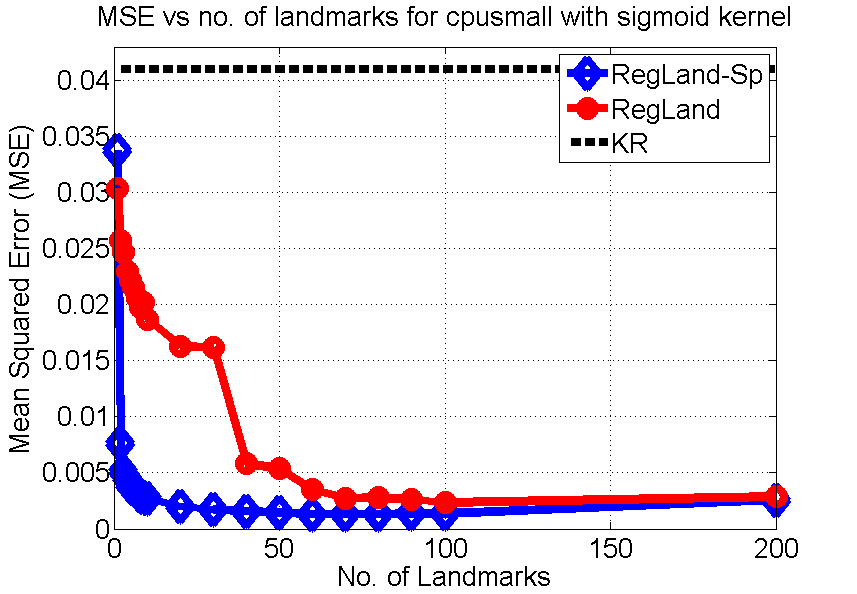}
		\hspace*{-2ex}
		\includegraphics[width=0.35\textwidth, angle=0]{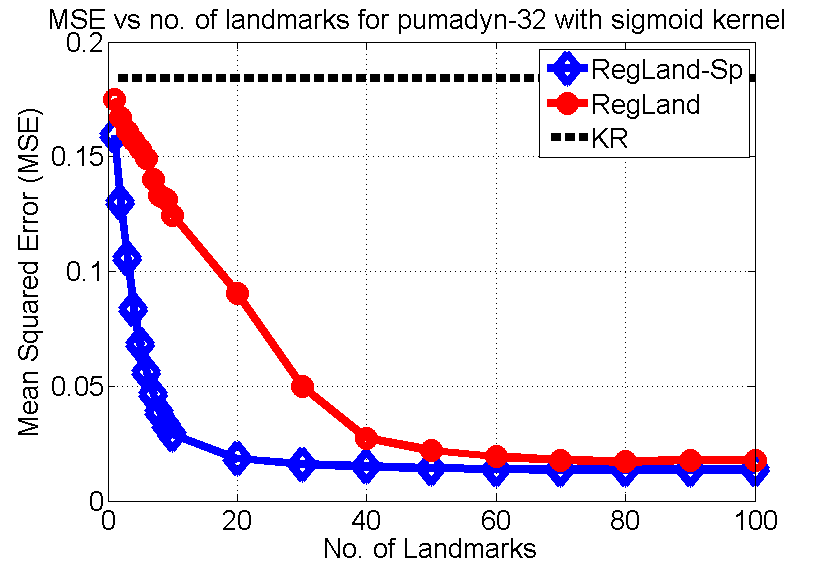}
	}\\\vspace*{-2ex}
	\subfloat[Avg. absolute error for landmarking (ORLand) and kernel regression (KR) on ordinal regression datasets]{
		\label{fig:ordreg}
		\includegraphics[width=0.35\textwidth, angle=0]{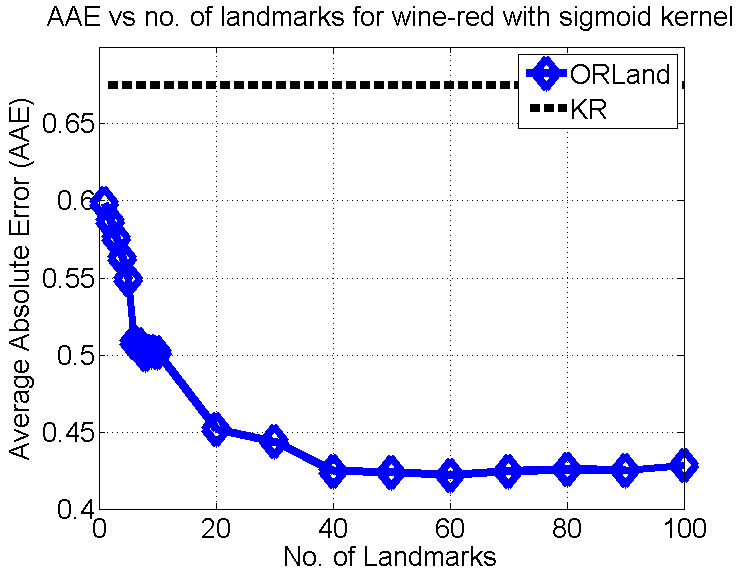}
		\hspace*{-2ex}
		\includegraphics[width=0.35\textwidth, angle=0]{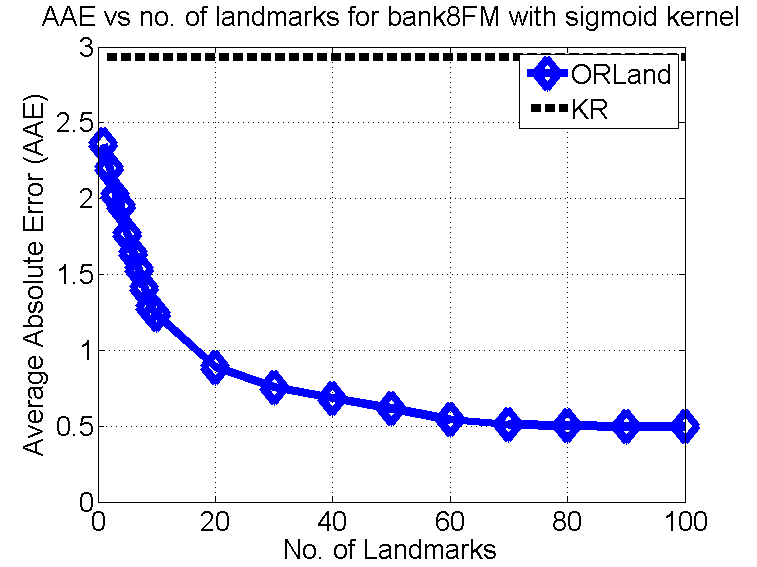}
		\hspace*{-2ex}
		\includegraphics[width=0.35\textwidth, angle=0]{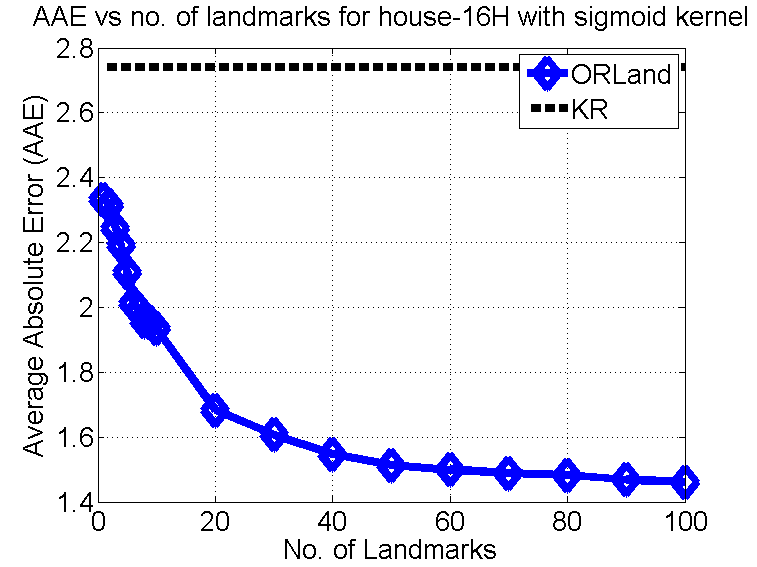}
	}
	\caption{Performance of landmarking algorithms with increasing number of landmarks on real-valued regression (Figure~\ref{fig:reg}) and ordinal regression (Figure~\ref{fig:ordreg}) datasets.}
	\label{fig:all_graphs}
\end{figure*}
\begin{table*}[t]
	\vspace*{-2ex}
	\hspace*{-11pt}
	\centering
	\subfloat[Mean squared error for real regression]{
		\label{tab:reg}
		\tiny
		\begin{tabular}{|m{0.6in}|m{0.35in}|m{0.35in}|m{0.35in}|m{0.35in}|}
			\hline
			\multirow{2}{*}{Datasets} & \multicolumn{2}{c|}{Sigmoid kernel} & \multicolumn{2}{c|}{Manhattan kernel} \\
			\cline{2-5}
			 & KR & Land-Sp & KR & Land-Sp\\ \hline
			 \texttt{Abalone} \cite{ucirep} \par $N = 4177$ \par$d = 8$ 		& 2.1e-002 \par (8.3e-004) & 6.2e-003 \par (8.4e-004) & 1.7e-002 \par (7.1e-004) & 6.0e-003 \par (3.7e-004) \\ \hline
			 \texttt{Bodyfat} \cite{statlibrep} \par $N = 252$ \par$d = 14$ 		& 4.6e-004 \par (6.5e-005) & 9.5e-005 \par (1.3e-004) & 3.9e-004 \par (2.2e-005) & 3.5e-005 \par (1.3e-005) \\ \hline
			 \texttt{CAHousing} \cite{statlibrep}\hspace*{-2ex} \par $N = 20640$ \par$d = 8$ 	& 5.9e-002 \par (2.3e-004) & 1.6e-002 \par (6.2e-004) & 5.8e-002 \par (1.9e-004) & 1.5e-002 \par (1.4e-004) \\ \hline
			 \texttt{CPUData} \cite{delverep} \par $N = 8192$ \par$d = 12$ 		& 4.1e-002 \par (1.6e-003) & 1.4e-003 \par (1.7e-004) & 4.3e-002 \par (1.6e-003) & 1.2e-003 \par (3.2e-005) \\ \hline
			 \texttt{PumaDyn-8} \cite{delverep}\hspace*{-2ex} \par $N = 8192$ \par$d = 8$ 	& 2.3e-001 \par (4.6e-003) & 1.4e-002 \par (4.5e-004) & 2.3e-001 \par (4.5e-003) & 1.4e-002 \par (4.8e-004) \\ \hline
			 \texttt{PumaDyn-32} \cite{delverep}\hspace*{-2ex} \par $N = 8192$ \par$d = 32$	& 1.8e-001 \par (3.6e-003) & 1.4e-002 \par (3.7e-004) & 1.8e-001 \par (3.6e-003) & 1.4e-002 \par (3.1e-004) \\ \hline
		\end{tabular}
	}
	\subfloat[Mean absolute error for ordinal regression]{
		\label{tab:ordreg}
		\tiny
		\begin{tabular}{|m{0.6in}|m{0.35in}|m{0.35in}|m{0.35in}|m{0.35in}|}
			\hline
			\multirow{2}{*}{Datasets} & \multicolumn{2}{c|}{Sigmoid kernel} & \multicolumn{2}{c|}{Manhattan kernel} \\
			\cline{2-5}
			 & KR & ORLand & KR & ORLand\\ \hline
			 \texttt{Wine-Red} \cite{ucirep} \par $N = 1599$ \par$d = 11$ 	& 6.8e-001 \par (2.8e-002) & 4.2e-001 \par (3.8e-002) & 6.7e-001 \par (3.0e-002) & 4.5e-001 \par (3.2e-002) \\ \hline
			 \texttt{Wine-White} \cite{ucirep}\hspace*{-2ex} \par $N = 4898$ \par$d = 11$ & 6.2e-001 \par (2.0e-002) & 8.9e-001 \par (8.5e-001) & 6.2e-001 \par (2.0e-002) & 4.9e-001 \par (1.5e-002) \\ \hline
			 \texttt{Bank-8} \cite{delverep} \par $N = 8192$ \par$d = 8$ 			& 2.9e+000 \par (6.2e-002) & 6.1e-001 \par (4.4e-002) & 2.7e+000 \par (6.6e-002) & 6.3e-001 \par (1.7e-002) \\ \hline
			 \texttt{Bank-32} \cite{delverep} \par $N = 8192$ \par$d = 32$ 		& 2.7e+000 \par (1.2e-001) & 1.6e+000 \par (2.3e-002) & 2.6e+000 \par (8.1e-002) & 1.6e+000 \par (9.4e-002) \\ \hline
			 \texttt{House-8} \cite{delverep} \par $N = 22784$ \par$d = 8$ 		& 2.8e+000 \par (9.3e-003) & 1.5e+000 \par (2.0e-002) & 2.7e+000 \par (1.0e-002) & 1.4e+000 \par (1.2e-002) \\ \hline
			 \texttt{House-16} \cite{delverep} \par $N = 22784$ \par$d = 16$ 	& 2.7e+000 \par (2.0e-002) & 1.5e+000 \par (1.0e-002) & 2.8e+000 \par (2.0e-002) & 1.4e+000 \par (2.3e-002) \\ \hline
		\end{tabular}
	}
	\\\vspace*{-1ex}
	\caption{Performance of landmarking-based algorithms (with 50 landmarks) vs. baseline kernel regression (KR). Values in parentheses indicate standard deviation values. Values in the first columns indicate dataset source (in parentheses), size (N) and dimensionality (d).}
	\label{tbl:uci}
\vspace*{-2ex}
\end{table*}

In this section we present an empirical evaluation of our learning models for the problems of real-valued regression and ordinal regression on benchmark datasets 
 taken from a variety of sources \cite{ucirep,statlibrep,delverep}. In all cases, we compare our algorithms against kernel regression (KR), a well known technique \cite{mlkr} for non-linear regression, whose predictor is of the form:
\[
f : \x \mapsto \frac{\sum_{\x_i \in \T} y(\x_i)K(\x,\x_i)}{\sum_{\x_i \in \T} K(\x,\x_i)}.
\]
where $\T$ is the training set. We selected KR as the baseline as it is a popular regression method that does not require similarity functions to be PSD. 
For ordinal regression problems, we rounded off the result of the KR predictor to get a discrete label. We implemented all our algorithms as well as the baseline KR method in Matlab. In all our experiments we report results across $5$ random splits on the (indefinite) Sigmoid: $K(\vecxy) = \tanh(a\ip{\vecx}{\vecy} + r)$ and Manhattan: $K(\vecxy) = -\norm{\vecx-\vecy}_1$ kernels. Following standard practice, we fixed $r = -1$ and $a = 1/d_{\text{orig}}$ for the Sigmoid kernel where $d_{\text{orig}}$ is the dimensionality of the dataset.


{\bf Real valued regression}:  For this experiment, we compare our methods (RegLand and RegLand-Sp) with the KR method. For RegLand, we constructed the landmarked space as specified in Algorithm~\ref{alg-learn-sim-gen} and learned a linear predictor using the LIBLINEAR package \cite{svr-lin} that minimizes $\epsilon$-insensitive loss. 
%
 In the second algorithm (RegLand-Sp), we used the sparse learning algorithm of \cite{acc-sparse} on the landmarked space to learn the best predictor for a given sparsity level. Due to its simplicity and good convergence properties, we implemented the \emph{Fully Corrective} version of the Forward Greedy Selection algorithm with squared loss as the surrogate.

We evaluated all methods using Mean Squared Error (MSE) on the test set. Figure~\ref{fig:reg} shows the MSE incurred by our methods along with reference values of accuracies obtained by KR as landmark sizes increase. 
 The plots clearly show that our methods incur significantly lesser error than KR. Moreover, RegLand-Sp learns more accurate predictors using the same number of landmarks. 
 For instance, when learning using the Sigmoid kernel on the \texttt{CPUData} dataset, at $20$ landmarks, RegLand is able to guarantee an MSE of $0.016$ whereas RegLand-Sp offers an MSE of less than $0.02$ ; MLKR is only able to guarantee an MSE rate of $0.04$ for this dataset. In Table~\ref{tab:reg}, we compare accuracies of the two algorithms when given $50$ landmark points with those of KR for the Sigmoid and Manhattan kernels. We find that in all cases, RegLand-Sp gives superior accuracies than KR. Moreover, the Manhattan kernel seems to match or outperform the Sigmoid kernel on all the datasets. 

{\bf Ordinal Regression}: Here, we compare our method with the baseline KR method on benchmark datasets. As mentioned in Section~\ref{sec:oreg}, our method uses the EXC formulation of \cite{ordreg-svm} along with landmarking scheme given in Algorithm~\ref{alg-learn-sim-gen}. We implemented a gradient descent-based solver (ORLand) to solve the primal formulation of EXC and used fixed 
equi-spaced thresholds instead of learning them as suggested by \cite{ordreg-svm}. Of the six datasets considered here, the two \texttt{Wine} datasets are ordinal regression datasets where the quality of the wine is to be predicted on a scale from $1$ to $10$. The remaining four datasets are regression datasets whose labels were subjected to equi-frequency binning to obtain ordinal regression datasets \cite{ordreg-svm}. We measured the average absolute error (AAE) for each method. Figure~\ref{fig:ordreg} compares ORLand with KR as the number of landmarks increases. 
Table~\ref{tab:ordreg} compares accuracies of ORLand for $50$ landmark points with those of KR for Sigmoid and Manhattan kernels. In almost all cases, ORLand gives a much better performance than KR. The Sigmoid kernel seems to outperform the Manhattan kernel on a couple of datasets.

We refer the reader to \ref{app:exps} for additional experimental results.


\vspace*{-1ex}
\section{Conclusion}
In this work we considered the general problem of supervised learning using non-PSD similarity functions. We provided a goodness criterion for similarity functions w.r.t. various learning tasks. This allowed us to construct efficient learning algorithms with provable generalization error bounds. At the same time, we were able to show, for each learning task, that our criterion is not too restrictive in that it admits all good PSD kernels. We then focused on the problem of identifying influential landmarks with the aim of learning sparse predictors. We presented a model that formalized the intuition that typically only a small fraction of landmarks is influential for a given learning problem. We adapted existing sparse vector recovery algorithms within our model to learn provably sparse predictors with bounded generalization error. Finally, we empirically evaluated our learning algorithms on benchmark regression and ordinal regression tasks. In all cases, our learning methods, especially the sparse recovery algorithm, consistently outperformed the kernel regression baseline.


An interesting direction for future research would be learning good similarity functions \'a la metric learning or kernel learning. It would also be interesting to conduct large scale experiments on real-world data such as social networks that naturally capture the notion of similarity amongst nodes. 

\subsubsection*{Acknowledgments}
P. K. is supported by a Microsoft Research India Ph.D. fellowship award. Part of this work was done while P. K. was an intern at Microsoft Research Labs India, Bangalore.

\clearpage
\newpage

\begin{small}
\bibliographystyle{unsrt}
\bibliography{refs}
\end{small}
\vspace*{10ex}
\appendix
\gdef\thesection{Appendix \Alph{section}}
\gdef\thesubsection{\Alph{section}.\arabic{subsection}}

\begin{center}
	{\LARGE \textbf{Supplementary Material}}
\end{center}

\vspace*{5ex}

Throughout this document, theorems and lemmata that were not originally proven as a part of this work cite, as a part of their statement, the work that originally presented the proof.

\section{Proofs of supplementary theorems}
\label{app:general}
In this section we give proofs for certain generic results that would be used in the utility and admissibility proofs. The first result, given as Lemma~\ref{landmark-guarantee}, allows us to analyze the landmarking step (Step~\ref{stp:land} of Algorithm~\ref{alg-learn-sim-gen}) and allows us to reduce the learning problem to that of learning a linear predictor over the landmarked space. The second result, given as Lemma~\ref{risk-bounds}, gives us a succinct re-statement of generalization error bounds proven in \cite{rademacher-averages} that would be used in proving utility bounds. The third result, given as Lemma~\ref{admissible-weights}, is a technical result that helps us prove admissibility bounds for our goodness definitions.

\begin{lem}[Landmarking approximation guarantee \cite{dissim-nips}]
\label{landmark-guarantee}
Given a similarity function $K$ over a domain $\X$ and a bounded function of the form $f(x) = \EE{\vecx' \sim \D}{w(\vecx')K(\vecx,\vecx')}$ for some bounded weight function $w : \X \rightarrow \bc{-B,B}$, for every $\epsilon,\delta > 0$ there exists a randomized map $\Psi : \X \rightarrow \R^d$ for $d = d\br{\epsilon,\delta}$ such that with probability at least $1-\delta$, there exists a linear operator $\tilde f$ over $\R^d$ such that $\EE{\vecx \sim \D}{\abs{\tilde f\br{\Psi\br{\vecx}} - f(\vecx)}} \leq \epsilon$.
\end{lem}
\begin{proof}
This result essentially allows us to project the learning problem into a Euclidean space where one can show, for the various learning problems considered here, that existing large margin techniques are applicable to solve the original problem. The result appeared in \cite{dissim-nips} and is presented here for completeness.

Sample $d$ \emph{landmark points} $\L = \bc{\vecx_1,\ldots,\vecx_d}$ from $\D$ and construct the map $\Psi_\L : \vecx \mapsto \frac{1}{\sqrt{d}}\br{K(\vecx,\vecx_1),\ldots,K(\vecx,\vecx_d)}$ and consider the linear operator $\tilde{f}$ over $\R^d$ defined as follows (in the following, we shall always omit the subscript $\L$ for clarity):
\[
\tilde{f} : \vecx \mapsto \frac{1}{d}\sum\limits_{i=1}^dw(\vecx_i)K(\vecx,\vecx_i) = \ip{\tilde\vecw}{\Psi(\vecx)}
\]
for $\vecw = \frac{1}{\sqrt{d}}\br{w(\vecx_1),\ldots,w(\vecx_d)} \in \R^d$. A standard Hoeffding-style argument shows that for $d = \O{\frac{B^2}{\epsilon^2}\log\frac{1}{\delta^2}} = \O{\frac{B^2}{\epsilon^2}\log\frac{1}{\delta}}$, $\tilde f$ gives a point wise approximation to $f$, i.e. for all $\vecx \in \X$, with probability greater than $1 - \delta^2$, we have $\abs{\tilde f(\Psi(\vecx)) - f(\vecx)} < \epsilon$.

Now call the event $\evbad{\vecx} := \abs{\tilde f(\Psi(\vecx)) - f(\vecx)} > \epsilon$. Thus we have for all $\vecx \in \X$, $\Prr{\tilde f}{\evbad{\vecx}} = \EE{\tilde f}{1_{\evbad{\vecx}}} < \delta^2$ (here the probabilities are being taken over the construction of $\tilde f$ i.e. the choice of the landmark points). Taking expectations over the entire domain, applying Fubini's theorem to switch expectations and applying Markov's inequality we get
\[
\Prr{\tilde f}{\Prr{\vecx \sim \D}{\evbad{\vecx}} > \delta} < \delta
\]
Thus with confidence $1 - \delta$ we have $\Prr{\vecx \sim \D}{\evbad{\vecx}} < \delta$ and thus $\EE{\vecx \sim \D}{\abs{\tilde f(\Psi(\vecx)) - f(\vecx)}} < (1-\delta)\epsilon + 2B\delta$ since $\underset{\vecx \in \X}{\sup}\abs{\tilde f(\Psi(\vecx))} = \underset{\vecx \in \X}{\sup}\abs{f(\vecx)} = B$. For $\delta < \frac{\epsilon}{B}$ we get $\EE{\vecx \sim \D}{\abs{\tilde f(\Psi(\vecx)) - f(\vecx)}} < 2 \epsilon$.
\end{proof}

\begin{lem}[Risk bounds for linear predictors \cite{rademacher-averages}]
\label{risk-bounds}
Consider a real-valued prediction problem $y$ over a domain $\X = \bc{\vecx : \norm{\vecx}_2 \leq C_X}$ and a linear learning model $\F : \bc{\vecx \mapsto \ip{\vecw}{\vecx} : \norm{\vecw}_2 \leq C_W}$ under some fixed loss function $\ell\br{\cdot,\cdot}$ that is $C_L$-Lipschitz in its second argument. For any $f \in \F$, let $\L_f = \EE{\vecx \sim \D}{\ell(f(\vecx),y(\vecx))}$ and $\hat \L_f^n$ be the empirical loss on a set of $n$ i.i.d. chosen points. Then we have, with probability greater than $\br{1 - \delta}$,
\[
\underset{f \in \F}{\sup}\br{\L_f - \hat \L_f^n} \leq 3C_LC_XC_W\sqrt{\frac{\log(1/\delta)}{n}}
\]
\end{lem}
\begin{proof}
There exist a few results that provide a unified analysis for the generalization properties of linear predictors \cite{rademacher-averages, covering-numbers}. However we use the heavy hammer of Rademacher average based analysis since it provides sharper bounds than covering number based analyses.

The result follows from imposing a squared $L_2$ regularization on the $\vecw$ vectors. Since the squared $L_2$ function is $2$-strongly convex with respect to the $L_2$ norm, using \cite[Theorem 1]{rademacher-averages}, we get a bound on the Rademacher complexity of the function class $\F$ as ${\cal R}_n\br{\F} \leq C_XC_W\sqrt{\frac{1}{n}}$. Next, using the Lipschitz properties of the loss function, a result from \cite{rademacher-risk-bounds} allows us to bound the excess error by $2C_L{\cal R}_n(\F) + C_LC_XC_W\sqrt{\frac{\log(1/\delta)}{2n}}$. The result then follows from simple manipulations.
\end{proof}

\begin{lem}[Admissible weight functions for PSD kernels \cite{kernel-sim-compare}]
\label{admissible-weights}
Consider a PSD kernel that is $\br{\epsilon_0,\gamma}$-good for a learning problem with respect to some convex loss function $\ell_K$. Then there exists a vector $\vecW' \in \H_K$ and a bounded weight function $w : \X \rightarrow \R$ such that $\EE{\vecx \sim \D}{\ell_K\br{\ip{\vecW'}{\Phi_K(\vecx)},y(\vecx)}} \leq \epsilon_0 + \frac{1}{2C\gamma^2}$ for some arbitrary positive constant $C$ and for all $\vecx \in \X$, we have $\EE{\vecx' \sim \D}{w(\vecx')K(\vecx,\vecx')} = \ip{\vecW'}{\Phi_K(\vecx)}$.
\end{lem}
\begin{proof}
Note that the $\br{\epsilon_0,\gamma}$-goodness of $K$ guarantees the existence of a weight vector $\vecW^\ast \in \H_K$ with small loss at large margin. Thus $\vecW'$ acts as a proxy for $\vecW^\ast$ providing bounded loss at unit margin but with the additional property of being functionally equivalent to a bounded weighted average of the kernel values as required by the definition of a good similarity function. This will help us prove admissibility results for our similarity learning models.

We start by proving the theorem for a discrete distribution - the generalization to non-discrete distributions will follow by using variational optimization techniques as discussed in \cite{kernel-sim-compare}. Consider a discrete learning problem with $\X = \bc{\vecx_1,\ldots,\vecx_n}$, corresponding distribution $\D = \bc{p_1,\ldots,p_n}$ and target $y = \bc{y_1,\ldots,y_n}$ such that $\sum p_i = 1$. Set up the following regularized ERM problem (albeit on the entire domain):
\begin{eqnarray*}
\underset{\vecW \in \H_K} \min	&& \frac{1}{2}\norm{\vecW}_{\H_K}^2 + C\sum\limits_{i=1}^n p_i \ell_K\br{\ip{\vecW}{\Phi_K(\vecx_i)},y_i}
\end{eqnarray*}
Let $\vecW'$ be the weight vector corresponding to the optima of the above problem. By the Representer Theorem (for example \cite{representer-theorem}), we can choose $\vecW' = \sum\alpha_i\Phi_K(\vecx_i)$ for some bounded $\alpha_i$ (the exact bounds on $\alpha_i$ are problem specific). By $\br{\epsilon_0,\gamma}$-goodness of $K$ we have
\begin{eqnarray*}
\frac{1}{2}\norm{\vecW'}_{\H_K}^2 + C\sum\limits_{i=1}^n p_i \ell_K\br{\ip{\vecW'}{\Phi_K(\vecx_i)},y_i} &\leq& \frac{1}{2}\norm{\frac{1}{\gamma}\vecW^\ast}_{\H_K}^2 + C\sum\limits_{i=1}^n p_i \ell_K\br{\frac{\ip{\vecW^\ast}{\Phi_K(\vecx_i)}}{\gamma},y_i} \\
&=& \frac{1}{2\gamma^2} + C \cdot \EE{\vecx \sim \D}{\ell_K\br{\frac{\ip{\vecW^\ast}{\Phi_K(\vecx)}}{\gamma},y(\vecx)}}\\
&\leq& \frac{1}{2\gamma^2} + C \epsilon_0
\end{eqnarray*}
Thus we have
\begin{eqnarray*}
\EE{\vecx \sim \D}{\ell_K\br{\ip{\vecW'}{\Phi_K(\vecx)},y(\vecx)}} &\leq& \frac{1}{2C}\norm{\vecW'}_{\H_K}^2 + \sum\limits_{i=1}^n p_i \ell_K\br{\ip{\vecW'}{\Phi_K(\vecx_i)},y_i}\\
&\leq& \epsilon_0 + \frac{1}{2C\gamma^2}
\end{eqnarray*}
which proves the first part of the claim. For the second part, set up a weight function $w_i = \frac{\alpha_i}{p_i}$. Then, for any $\vecx \in \X$ we have
\begin{eqnarray*}
\EE{\vecx' \sim \D}{w(\vecx')K(\vecx,\vecx')} &=& \sum\limits_{i=1}^n p_i w_i K(\vecx,\vecx_i)= \sum\limits_{i=1}^n p_i \frac{\alpha_i}{p_i} K(\vecx,\vecx_i)\\
		&=& \sum\limits_{i=1}^n \alpha_i\ip{\Phi_K(\vecx)}{\Phi_K(\vecx_i)} = \ip{\vecW'}{\Phi_K(\vecx)}
\end{eqnarray*}
The weight function is bounded since the $\alpha_i$ are bounded and, this being a discrete learning problem, cannot have vanishing probability masses $p_i$ (actually, in the cases we shall consider, the $\alpha_i$ will itself contain a $p_i$ term that will subsequently get cancelled). For non-discrete cases, variational techniques give us similar results.
\end{proof}

\section{Justifying Double-dipping}
\label{app:double-dipping}
All our analyses (as well as the analyses presented in \cite{learn-sim, learn-dissim, dissim-nips}) use some data as landmark points and then require a fresh batch of training points to learn a classifier on the landmarked space. In practice, however, it might be useful to reuse training data to act as landmark points as well. This is especially true of \cite{learn-dissim, dissim-nips} who require labeled landmarks. We give below, generalization bounds for similarity-based learning algorithms that indulge in such ``double dipping''. The argument uses a technique outlined in \cite{double-dipping} and falls within the Rademacher-average based uniform convergence guarantees used elsewhere in the paper. We present a generic argument that, in a manner similar to Lemma~\ref{risk-bounds}, can be specialized to the various learning problems considered in this paper.

To make the presentation easier we set up some notation. For any predictor $f$, let $\L_f = \EE{\vecx \sim \D}{\ell(f(\vecx),y(\vecx))}$ and for any training set $S$ of size $n$, let $\hat \L_f^S = \frac{1}{n}\sum_{\vecx_i \in S}\ell(f(\vecx_i),y(\vecx_i))$. For any landmark set $S = \br{\vecx_1,\ldots,\vecx_n}$, we let $\Psi_S : \vecx \mapsto \br{K(\vecx,\vecx_1),\ldots,K(\vecx,\vecx_n)}$. For any weight vector $\vecw \in \R^n, \norm{\vecw}_\infty \leq B$ in the landmarked space, denote the predictor $f_{\br{S,\vecw}} := \frac{1}{n}\ip{\vecw}{\Psi_S(\vecx)} = \vecx \mapsto \frac{1}{n}\sum\limits_{i=1}^n\vecw_iK(\vecx,\vecx_i)$. Also let $\F_S := \bc{\vecx \mapsto \frac{1}{n}\ip{\vecw}{\Psi_S(\vecx)}} = \bc{f_{\br{S,\vecw}} : \vecw \in \R^n, \norm{\vecw}_\infty \leq B}$.

We note that the embedding defined above is ``stable'' in the sense that changing a single landmark does not change the embedding too much with respect to bounded predictors. More formally, for any set of $n$ points $S = \br{\vecx_1,\ldots,\vecx_n}$, define $g(S) := \underset{f \in \F_S}{\sup}\bc{\L_{f} - \hat\L_{f}^S}$. Let $S^i$ be another set of $n$ points that (arbitrarily) differs from $S$ just at the $i^{\text{th}}$ point and coincides with $S$ on the rest. Then we have, for any fixed $\vecw$ of bounded $L_\infty$ norm (i.e. $\norm{\vecw}_\infty \leq B$) and bounded similarity function (i.e. $K(\vecxy) \leq 1$),
\begin{eqnarray*}
\underset{\vecx}{\sup}\bc{\abs{f_{\br{S,\vecw}}(\vecx) - f_{\br{{S^i},\vecw}}(\vecx)}} &=& \underset{\vecx}{\sup}\bc{\abs{\frac{1}{n}\sum\limits_{j=1}^n\vecw_jK(\vecx,\vecx_j) - \frac{1}{n}\sum\limits_{i=1}^n\vecw_jK(\vecx,\vecx_j')}}\\
								&=& \underset{\vecx}{\sup}\bc{\abs{\frac{1}{n}\vecw_i\br{K(\vecx,\vecx_i) - K(\vecx,\vecx_i')}}}\\
								&\leq& \frac{2B}{n}
\end{eqnarray*}

Note that, although \cite{dissim-nips} uses pairs of labeled points to define the embedding, the following argument can easily be extended to incorporate this since the embedding is identical to the embedding $\Psi_S$ described above with respect to being ``stable''. In fact this analysis holds for any stable embedding defined using training points.

Our argument proceeds by showing that with high probability (over choice of the set $S$) we have
\[
\underset{\vecw}\sup\bc{\L_{f_{\br{S,\vecw}}} - \hat\L_{f_{\br{S,\vecw}}}^S} \leq \epsilon
\]
By the definition of $\F_S$, the above requirement translates to showing that with high probability,
\[
\underset{f \in \F_S}\sup\bc{\L_{f} - \hat\L_{f}^S} \leq \epsilon
\]
which highlights the fact that we are dealing with a problem of sample dependent hypothesis spaces\footnote{We were not able to find any written manuscript detailing the argument of \cite{double-dipping}. However the argument itself is fairly generic in allowing one to prove generalization bounds for sample dependent hypothesis spaces.}.
Note that this exactly captures the double dipping procedure of reusing training points as landmark points. Such a result would be useful as follows: using Lemma~\ref{landmark-guarantee} and task specific guarantees (outlined in detail in the subsequent sections), we have, with high probability, the existence of a good predictor in the landmarked space of a randomly chosen landmark set $S$ i.e. with very high probability over choice of $S$, we have $\underset{f \in \F_S}\inf\bc{\L_{f}} \leq \epsilon_0$. Let this be achieved by the predictor $f^\ast$. Using the uniform convergence guarantee above we get $\hat\L_{f^\ast}^S \leq \epsilon_0 + \epsilon$ (with some loss of confidence due to application of a union bound).

Now consider the predictor $\hat f := \underset{f \in \F_S}\inf\bc{\hat\L_{f}^S}$. Clearly $\hat\L_{\hat f}^S \leq \hat\L_{f^\ast}^S \leq \epsilon_0 + \epsilon$. Invoking the uniform convergence bound yet again shows us that
\[
\L_{\hat f} \leq \hat\L_{\hat f}^S + \underset{f \in \F_S}\sup\bc{\L_{f} - \hat\L_{f}^S} \leq \epsilon_0 + 2\epsilon
\]
Note that we incur some more loss of confidence due to another application of the union bound. This tells us that with high probability, a predictor learned by choosing a random landmark set and training on the landmark set itself would yield a good predictor.

We will proceed via a vanilla uniform convergence argument involving symmetrization and an application of the McDiarmid's inequality (stated below). However, proving the stability prerequisite for the application of the McDiarmid's inequality shall require use of the stability of both the predictor $f_{\br{S,\vecw}}$ as well as the embedding $\Psi_S$. Let the loss function $\ell$ be $C_L$-Lipschitz in its first argument.

\begin{thm}[McDiarmid's inequality \cite{mcdiarmid}]
Let $X_1,\ldots,X_n$ be independent random variables taking values in some set $\X$. Further, let $f : \X^n \rightarrow \R$ be a function of $n$ variables that satisfies, for all $i \in [n]$ and all $x_1,\ldots,x_n,x_i' \in \X$,
\[
\abs{f\br{x_1,\ldots,x_i,\ldots,x_n} - f\br{x_1,\ldots,x_i',\ldots,x_n}} \leq c_i
\]
then for all $\epsilon > 0$, we have
\[
\Pr{f - \E{f} > \epsilon} \leq \exp\br{\frac{-2 \epsilon^2}{\sum_{i=1}^nc_i^2}}
\]
\end{thm}

We shall invoke the McDiarmid's inequality on the function $g(S) := \underset{f \in \F_S}{\sup}\bc{\L_{f} - \hat\L_{f}^S}$ with $S = \br{\vecx_1,\ldots,\vecx_n}$ being the random variables in question. To do so we first prove the stability of the function $g(S)$ with respect to its variables and then bound the value of $\EE{S}{g(S)}$.

\begin{thm}
For any $S, S^i$, we have $\abs{g(S) - g(S^i)} \leq \frac{6BC_L}{n}$.
\end{thm}
\begin{proof}
We have
\begin{eqnarray*}
g(S) &=& \underset{f \in \F_S}{\sup}\bc{\L_{f} - \hat\L_{f}^S}\\
		 &=& \underset{f \in \F_S}{\sup}\bc{\L_{f} - \hat\L_{f}^S - \hat\L_{f}^{S^i} + \hat\L_{f}^{S^i}}\\
		 &\leq& \underset{f \in \F_S}{\sup}\bc{\L_{f}  - \hat\L_{f}^{S^i}} + \underset{f \in \F_S}{\sup}\bc{\hat\L_{f}^S - \hat\L_{f}^{S^i}}\\
		 &\leq& \underset{f \in \F_S}{\sup}\bc{\L_{f}  - \hat\L_{f}^{S^i}} + \frac{2BC_L}{n}
\end{eqnarray*}
where in the fourth step we have used the fact that the loss function is Lipschitz and the embedding function $\Psi_S$ is bounded. We also have
\begin{eqnarray*}
\underset{f \in \F_S}{\sup}\bc{\L_{f}  - \hat\L_{f}^{S^i}} &=& \underset{\vecw}\sup\bc{\L_{f_{\br{S,\vecw}}} - \hat\L_{f_{\br{S,\vecw}}}^{S^i}}\\
					&=& \underset{\vecw}\sup\bc{\L_{f_{\br{S,\vecw}}} - \L_{f_{\br{{S^i},\vecw}}} + \L_{f_{\br{{S^i},\vecw}}} - \hat\L_{f_{\br{{S^i},\vecw}}}^{S^i} + \hat\L_{f_{\br{{S^i},\vecw}}}^{S^i} - \hat\L_{f_{\br{S,\vecw}}}^{S^i}}\\
					&\leq& \underset{\vecw}\sup\bc{\L_{f_{\br{{S^i},\vecw}}} - \hat\L_{f_{\br{{S^i},\vecw}}}^{S^i}} + \underset{\vecw}\sup\bc{\L_{f_{\br{S,\vecw}}} - \L_{f_{\br{{S^i},\vecw}}}}\\
					&& + \underset{\vecw}\sup\bc{\hat\L_{f_{\br{{S^i},\vecw}}}^{S^i} - \hat\L_{f_{\br{S,\vecw}}}^{S^i}}\\
					&\leq& \underset{\vecw}\sup\bc{\L_{f_{\br{{S^i},\vecw}}} - \hat\L_{f_{\br{{S^i},\vecw}}}^{S^i}} + \frac{2BC_L}{n} + \frac{2BC_L}{n}\\
					&=& \underset{f \in \F_{S^i}}{\sup}\bc{\L_{f} - \hat\L_{f}^{S^i}} + \frac{4BC_L}{n}\\
					&=& g(S^i) + \frac{4BC_L}{n}
\end{eqnarray*}
where in the fourth step we have used the stability of the embedding function and that the loss function is $C_L$-Lipschitz in its first argument so that for all $\vecx$ we have $\displaystyle \abs{\ell\br{f_{\br{S,\vecw}}(\vecx),y(\vecx)} - \ell\br{f_{\br{{S^i},\vecw}}(\vecx),y(\vecx)}} \leq \frac{2BC_L}{n}$ which holds in expectation over any (empirical) distribution as well. Putting the two inequalities together gives us $\displaystyle g(S) \leq g(S^i) + \frac{6BC_L}{n}$. Similarly we also have $\displaystyle g(S^i) \leq g(S) + \frac{6BC_L}{n}$ which gives us the result.
\end{proof}

We now have that the function $g(S)$ is $\displaystyle \O{\frac{1}{n}}$-stable with respect to each of its inputs. We now move on to bound its expectation. For any function class $\F$ we define its \emph{empirical Rademacher average} as follows
\[
\hat\RR_n(\F) := \EE{\sigma}{\left.\underset{f \in \F}{\sup}\bc{\frac{1}{n}\sum_{\vecx_i \in S}\sigma_if(\vecx_i)}\right|S}
\]
Also let $\F := \bc{\vecx \mapsto \ip{\vecw}{\vecx} : \norm{\vecw}_2 \leq B}$ and $\X := \bc{\vecx : \norm{\vecx}_2 \leq 1}$.

\begin{thm}
$\EE{S}{\underset{f \in \F_S}{\sup}\bc{\L_{f} - \hat\L_{f}^S}} \leq 2BC_L\sqrt\frac{1}{n}$
\end{thm}
\begin{proof}
We have
\begin{eqnarray*}
\EE{S}{\underset{f \in \F_S}{\sup}\bc{\L_{f} - \hat\L_{f}^S}} &=& \EE{S}{\underset{f \in \F_S}{\sup}\bc{\EE{S'}{\hat\L_{f}^{S'}} - \hat\L_{f}^S}}\\
			&\leq& \EE{S,S'}{\underset{f \in \F_S}{\sup}\bc{\hat\L_{f}^{S'} - \hat\L_{f}^S}}\\
			&\leq& \EE{S,S'}{\underset{f \in \F_{S \cup S'}}{\sup}\bc{\hat\L_{f}^{S'} - \hat\L_{f}^S}}\\
			&=& \EE{S,S',\sigma}{\underset{f \in \F_{S \cup S'}}{\sup}\bc{\frac{1}{n}\sum_{\vecx_i \in S,\vecx_i' \in S'}\sigma_i\br{\ell(f(\vecx_i'),y(\vecx_i')) - \ell(f(\vecx_i),y(\vecx_i))}}}\\
			&\leq& 2\EE{S,S',\sigma}{\underset{f \in \F_{S \cup S'}}{\sup}\bc{\frac{1}{n}\sum_{\vecx_i \in S}\sigma_i\ell(f(\vecx_i),y(\vecx_i))}}\\
			&=& 2\EE{S,S',\sigma}{\underset{\vecw}{\sup}\bc{\frac{1}{n}\sum_{\vecx_i \in S}\sigma_i\ell(f_{\br{{S\cup S'},\vecw}}(\vecx_i),y(\vecx_i))}}\\
			&\leq& 2\EE{S,S',\sigma}{\underset{f \in \F}{\sup}\bc{\frac{1}{n}\sum_{\vecx_i \in S}\sigma_i\ell(f(\vecx_i),y(\vecx_i))}}\\
			&=& 2\ \EE{S}{\hat\RR_n(\ell\circ\F)} \leq 2C_L\ \EE{S}{\hat\RR_n(\F)} \leq 2BC_L\sqrt\frac{1}{n}\\
\end{eqnarray*}
where in the third step we have used the fact that $\F_S \supseteq \F_{S'}$ if $S \supseteq S'$ (this is the monotonicity requirement in \cite{double-dipping}). Note that this is essential to introduce symmetry so that Rademacher variables can be introduced in the next (symmetrization) step. In the seventh step, we have used the fact that for every $S$ such that $\abs{S} = n$ and $\vecw \in \R^n$ such that $\norm{\vecw}_\infty \leq B$, there exists a function $f \in \F$ such that for all $\vecx$, there exists a $\vecx' \in \X$ such that $f_{\br{{S},\vecw}}(\vecx) = f(\vecx')$. In the last step we have used a result from \cite{rad-lip} which allows calculation of Rademacher averages for composition classes and an intermediate result from the proof of Lemma~\ref{risk-bounds} which gives us Rademacher averages for the function class $\F$.
\end{proof}
Thus, by an application of McDiarmid's inequality we have, with probability $\br{1-\delta}$ over choice of the landmark (training) set,
\[
\underset{f \in \F_S}\sup\bc{\L_{f} - \hat\L_{f}^S} \leq \E{\underset{f \in \F_S}{\sup}\bc{\L_{f} - \hat\L_{f}^S}} + 6BC_L\sqrt{\frac{\log 1/\delta}{2n}} \leq 4BC_L\sqrt{\frac{\log 1/\delta}{n}}
\]
which concludes our argument justifying double dipping.
\section{Regression with Similarity Functions}
\label{sec:app_reg}
In this section we give proofs of utility and admissibility results for our similarity based learning model for real-valued regression tasks.

\subsection{Proof of Theorem~\ref{good-sim-guarantee}}
First of all, we use Lemma~\ref{landmark-guarantee} to project onto a $d$ dimensional space where there exists a linear predictor $\tilde f : \vecx \mapsto \ip{\vecw}{\vecx}$ such that $\EE{\vecx \sim \D}{\abs{\tilde f\br{\Psi\br{\vecx}} - f(\vecx)}} \leq 2\epsilon_1$. Note that $\norm{\vecw}_2 \leq B$ and $\underset{\vecx \in \X}{\sup}\bc{\norm{\Psi(\vecx)}} \leq 1$ by construction. We will now show that $\tilde f$ has bounded $\epsilon$-insensitive loss.

%
\begin{eqnarray*}
\EE{\vecx \sim \D}{\eild{\tilde f\br{\Psi\br{\vecx}}, y(\vecx)}} &=& \EE{\vecx \sim \D}{\eild{f(\vecx), y(\vecx)}} + \EE{\vecx \sim \D}{\eild{\tilde f\br{\Psi\br{\vecx}}, y(\vecx)} - \eild{f(\vecx), y(\vecx)}}\\
                                                    &\leq& \epsilon_0 + \EE{\vecx \sim \D}{\eild{\tilde f\br{\Psi\br{\vecx}}, y(\vecx)} - \eild{f(\vecx), y(\vecx)}}\\
                                                    &\leq& \epsilon_0 + \EE{\vecx \sim \D}{\abs{\tilde f\br{\Psi(\vecx)} - f(\vecx)}}\\
                                                    &\leq& \epsilon_0 + 2 \epsilon_1\\
\end{eqnarray*}
where in the second step we have used the goodness properties of $K$, in the third step we used the fact that the $\epsilon$-insensitive loss function is $1$-Lipschitz in its first argument. Note that $\norm{\vecw} \approx \EE{\vecx \sim \D}{w^2(\vecx)}$ with high probability and if $\EE{\vecx \sim \D}{w^2(\vecx)} \ll B$ then we get a much better bound on the norm of $\vecw$. The excess loss incurred due to this landmarking step is, with probability $1 - \delta$, at most $32B\sqrt{\frac{\log(1/\delta)}{d}}$.

Now consider the following regularized ERM problem on $n$ i.i.d. sample points:
\[
\hat\vecw = \underset{\vecw : \norm{\vecw}_2 \leq B}{\arg\min} \frac{1}{n}\sum\limits_{i=1}^n\eild{\ip{\vecw}{\Psi(\vecx_i)},y(\vecx_i)}
\]
The final output of our learning algorithm shall be $\vecx \mapsto \ip{\hat\vecw}{\Psi(\vecx)}$. Here we have $C_X = 1$, $C_L = 1$ since $\eild{\cdot}$ is $1$-Lipschitz and $C_W = B$. Thus by Lemma~\ref{risk-bounds}, we get that the excess loss incurred due to this regularized ERM step is at most $3B\sqrt{\frac{\log{1/\delta}}{n}}$.

Since the $\epsilon$-insensitive loss is related to the absolute error by $\abs{x} \leq \eild{x} + \epsilon$ we have the total error (with respect to absolute loss) being incurred by our predictor to be, with probability at least $1 - 2\delta$, at most
\[
\epsilon_0 + 32B\sqrt{\frac{\log(1/\delta)}{d}} + 3B\sqrt{\frac{\log{1/\delta}}{n}} + \epsilon
\]

Taking $d = \O{\frac{B^2}{\epsilon_1^2}\log{\frac{1}{\delta}}}$ unlabeled landmarks and $n = \O{\frac{B^2}{\epsilon_1^2}\log{\frac{1}{\delta}}}$ labeled training points gives us our desired result.
%

\subsection{Proof of Theorem~\ref{reg-psd-sim-good}}
We prove the two parts of the result separately.

{\bf Part 1: Admissibility}: Using Lemma~\ref{admissible-weights} it is possible to obtain a vector $\vecW' = \sum\limits_{i=1}^n(\alpha_i - \alpha_i^\ast)\Phi_K(\vecx_i) \in \H_K$ with small loss such that $0 \leq \alpha_i,\alpha_i^\ast \leq p_iC$ and $\alpha_i \alpha_i^\ast  = 0$ (these inequalities are a consequence of applying the KKT conditions). This allows us to construct as weight function $w_i = \frac{\alpha_i - \alpha_i^\ast}{p_i}$ such that $\abs{w_i} \leq C$ and $\EE{\vecx' \sim \D}{w(\vecx')K(\vecx,\vecx')} = \ip{\vecW'}{\Phi_K(\vecx)}$ for all $\vecx \in \X$.

Thus we have $\EE{\vecx \sim \D}{\eild{\EE{\vecx' \sim \D}{w(\vecx')K(\vecx,\vecx')},y(\vecx)}} = \EE{\vecx \sim \D}{\eild{\ip{\vecW'}{\Phi_K(\vecx)},y(\vecx)}} \leq \frac{1}{2C\gamma^2} + \epsilon_0$. Setting $C = \frac{1}{2 \epsilon_1 \gamma^2}$ gives us our result.

We can use variational techniques to extend this to non-discrete distributions as well.


{\bf Part 2: Tightness}: The tight example that we provide is an adaptation of the example given for large margin classification in \cite{kernel-sim-compare}. However, our analysis differs from that of \cite{kernel-sim-compare}, partly necessitated by our choice of loss function.

Consider the following regression problem: $\X = \bc{\vecx_1,\vecx_2,\vecx_3,\vecx_4} \subset \R^3$, $\D = \bc{\frac{1}{2}- \epsilon, \epsilon, \epsilon, \frac{1}{2} - \epsilon}$, $y = \bc{+1,+1,-1,-1}$ 
\begin{eqnarray*}
\vecx_1 &=& \br{\gamma,\gamma,\sqrt{1 - 2\gamma^2}}\\
\vecx_2 &=& \br{\gamma,-\gamma,\sqrt{1 - 2\gamma^2}}\\
\vecx_3 &=& \br{-\gamma,\gamma,\sqrt{1 - 2\gamma^2}}\\
\vecx_4 &=& \br{-\gamma,-\gamma,\sqrt{1 - 2\gamma^2}}
\end{eqnarray*}
Clearly the vector $\vecw = \br{1,0,0}$ yields a predictor $y'$ with no $\epsilon$-insensitive loss for $\epsilon = 0$ (i.e. $\EE{\vecx \sim \D}{\eil{0}{y(\vecx) - y'(\vecx)}} = 0$) at margin $\gamma$. Thus the native inner product $\ip{\cdot}{\cdot}$ on $\R^3$ is a $\br{0,\gamma}$-good kernel for this particular regression problem.

Now consider any bounded weighing function on $\X$, $w = \bc{w_1,w_2,w_3,w_4}$ and analyze the effectiveness of $\ip{\cdot}{\cdot}$ as a similarity function. The output $\tilde y$ of the resulting predictor on the different points is given by $\tilde y_i = \sum\limits_{j=1}^4p_jw_j\ip{\vecx_i}{\vecx_j}$.

In particular, consider the output on the \emph{heavy} points $\vecx_1$ and $\vecx_4$ (note that the analysis in \cite{kernel-sim-compare} considers the \emph{light} points $\vecx_2$ and $\vecx_3$ instead). We have
\begin{eqnarray*}
\tilde y_1 &=& \br{\frac{1}{2} - \epsilon}w_1 + \epsilon \br{1- 2\gamma^2}\br{w_2 + w_3} + \br{\frac{1}{2} - \epsilon}w_4\br{1 - 4\gamma^2} = a + \br{\frac{1}{2} - \epsilon}\br{w_1 + bw_4}\\\
\tilde y_4 &=& \br{\frac{1}{2} - \epsilon}w_1\br{1 - 4\gamma^2} + \epsilon \br{1- 2\gamma^2}\br{w_2 + w_3} + \br{\frac{1}{2} - \epsilon}w_4 = a + \br{\frac{1}{2} - \epsilon}\br{bw_1 + w_4}
\end{eqnarray*}
for $a = \epsilon \br{1- 2\gamma^2}\br{w_2 + w_3}, b = \br{1 - 4\gamma^2}$. The main idea behind this choice is that the difference in the value of the predictor on these points is only due to the values of $w_1$ and $w_4$. Since the true values at these points are very different, this should force $w_1$ and $w_4$ to take large values unless a large error is incurred. To formalize this argument we lower bound the expected $\eil{0}{\cdot}$ loss of this predictor by the loss incurred on these heavy points.
\begin{eqnarray*}
\EE{\vecx \sim \D}{\eil{0}{y(\vecx) - \tilde y(\vecx)}} &\geq& \br{\frac{1}{2} - \epsilon}\br{\eil{0}{y(\vecx_1) - \tilde y(\vecx_1)} + \eil{0}{y(\vecx_4) - \tilde y(\vecx_4)}}\\
 &=& \br{\frac{1}{2} - \epsilon}\br{\abs{1 - \tilde y(\vecx_1)} + \abs{-1 - \tilde y(\vecx_4)}}\\
 &\geq& \br{\frac{1}{2} - \epsilon}\br{2 - \tilde y(\vecx_1) + \tilde y(\vecx_4)}\\
 &=& \br{\frac{1}{2} - \epsilon}\br{2 - \br{\frac{1}{2} - \epsilon}\br{1-b}\br{w_4 - w_1}}\\
 &=& \br{\frac{1}{2} - \epsilon}\br{2 - \br{\frac{1}{2} - \epsilon}\br{4\gamma^2}\br{w_4 - w_1}}
\end{eqnarray*}
where in the second step we use the fact that $\eil{0}{x} = \abs{x}$ and in the third step we used the fact that $\abs{a}+\abs{b} \geq a - b$. Thus, in order to have expected error at most $\epsilon_1$, we require
\[
w_4 - w_1 \geq \frac{1}{4\gamma^2}\br{2 - \frac{\epsilon_1}{\frac{1}{2} - \epsilon}}\frac{1}{\frac{1}{2} - \epsilon} = \frac{1}{4 \epsilon_1 \gamma^2}
\]
for the setting $\epsilon = \frac{1}{2} - \epsilon_1$. Thus we have $\abs{w_1} + \abs{w_4} \geq w_4 - w_1 \geq \frac{1}{4 \epsilon_1 \gamma^2}$ which implies $\max\br{\abs{w_1},\abs{w_4}} \geq \frac{1}{8 \epsilon_1 \gamma^2}$ which proves the result.

\section{Sparse Regression with Similarity functions}
\label{app:sp_reg}
Our utility proof proceeds in three steps. In the first step we project our learning problem, via the landmarking step given in Step~\ref{stp:land} of Algorithm~\ref{alg-learn-sim-gen}, to a linear landmarked space and show that the landmarked space admits a sparse linear predictor with bounded $\epsilon$-insensitive loss. This is formalized in Theorem~\ref{sparse-landmark-guarantee} which we restate for convenience.

\begin{thm}[Theorem~\ref{sparse-landmark-guarantee} restated]
Given a similarity function that is $\br{\epsilon_0,B,\tau}$-good for a regression problem, there exists a randomized map $\Psi : \X \rightarrow \R^d$ for $d = \O{\frac{B^2}{\tau\epsilon_1^2}\log\frac{1}{\delta}}$ such that with probability at least $1-\delta$, there exists a linear operator $\tilde f : \vecx \mapsto \ip{\vecw}{\vecx}$ over $\R^d$ such that $\norm{\vecw}_1 \leq B$ with $\epsilon$-insensitive loss bounded by $\epsilon_0 + \epsilon_1$. Moreover, with the same confidence we have $\norm{\vecw}_0 \leq \frac{3d\tau}{2}$.
\end{thm}
\begin{proof}
The proof of this theorem essentially parallels that of \cite[Theorem 8]{learn-sim-improv} but diverges later since the aim there is to preserve margin violations whereas we wish to preserve loss under the absolute loss function. Sample $d$ \emph{landmark points} $\L = \bc{\vecx_1,\ldots,\vecx_d}$ from the distribution $\D$ and construct the map $\Psi_\L : \vecx \mapsto \br{K(\vecx,\vecx_1),\ldots,K(\vecx,\vecx_d)}$ and consider the linear operator $\tilde{f} : \vecx \mapsto \ip{\vecw}{\vecx}$ with $\vecw_i = \frac{w(\vecx_i)R(\vecx_i)}{d_{\text{info}}}$ where $d_{\text{info}} = \sum\limits_{i=1}^dR(\vecx_i)$ is the number of informative landmarks. In the following we will refer to $\tilde f$ and $\vecw$ interchangeably. This ensures that $\norm{\tilde f}_1 := \norm{\vecw}_1 \leq B$. Note that we have chosen an $L_1$ normalized weight vector instead of an $L_2$ normalized one like we had in Lemma~\ref{landmark-guarantee}. This is due to a subsequent use of sparsity promoting regularizers whose analysis requires the existence of bounded $L_1$ norm predictors.

Using the arguments given for Lemma~\ref{landmark-guarantee} and Theorem~\ref{good-sim-guarantee}, we can show that if $d_{\text{info}} = \Om{\frac{B^2}{\epsilon_1^2}\log\frac{1}{\delta}}$ (i.e. if we have collected enough informative landmarks), then we are done. However, the Chernoff bound (lower tail) tells us that for $d = \Om{\frac{B^2}{\tau\epsilon_1^2}\log\frac{1}{\delta}}$, this will happen with probability $1 - \delta$. Moreover, the Chernoff bound (upper tail) tells us that, simultaneously we will also have $d_{\text{inf}} \leq \frac{3d\tau}{2}$. Together these prove the claim.
\end{proof}

Note that the number of informative landmarks required is, up to constant factors, the same as the number required in Theorem~\ref{good-sim-guarantee}. However, we see that in order to get these many informative landmarks, we have to sample a much larger number number of landmarks.  In the following, we shall see how to extract a sparse predictor in the landmarked space with good generalization properties. The following analysis shall assume the the existence of a good predictor on the landmarked space and hence all subsequent results shall be conditioned on the guarantees given by Theorem~\ref{sparse-landmark-guarantee}.

\begin{figure*}[t]
	\vspace*{-10pt}
	\begin{minipage}{\textwidth}
		\begin{algorithm}[H]
			\caption{\small Sparse regression \cite{acc-sparse}}
			\label{alg-sparse-reg}
			\begin{algorithmic}[1]
				\small{
					\REQUIRE A $\beta$-smooth loss function $\ell(\cdot,\cdot)$, regularization parameter $C_W$ used in Equation~\ref{prog-sparse}, error tolerance $\epsilon$
					\ENSURE A sparse predictor $\hat\vecw$ with bounded loss
					\STATE $k \leftarrow \left\lceil\frac{8C_W^2}{\epsilon^2}\right\rceil$, $\vecw^{(0)} = \veczero$
					\FOR{$t = 1$ \TO $k$}
						\STATE $\vectheta^{(t)}\leftarrow \nabla_\vecw\RR(\vecw^{(t)}) = \EE{\vecx \sim \D}{\frac{\partial}{\partial\vecw}\ell\br{\ip{\vecw^{(t)}}{\vecx},y(\vecx)}}$
						\STATE $r_t = \underset{j \in d}{\arg\max}\abs{\vectheta^{(t)}_j}$
						\STATE $\delta_t = \ip{\vectheta^{(t)}}{\vecw^{(t)}}+C_W\norm{\vectheta^{(t)}}_\infty$
						\STATE $\eta_t = \min\bc{1,\frac{\delta_t}{4C_W^2\beta}}$
						\STATE $\vecw^{(t+1)} \leftarrow \br{1 - \eta_t}\vecw^{(t)} + \eta_t\sign\br{-\vectheta^{(t)}_{r_t}}C_W\vec{e}^{r_t}$
						\IF{$\delta_t \leq \epsilon$}
							\RETURN{$\vecw^{(t)}$}
						\ENDIF
					\ENDFOR
					\RETURN{$\vecw^{(k)}$}
				}
			\end{algorithmic}
		\end{algorithm}
	\end{minipage}
	\vspace*{-10pt}
\end{figure*}

\subsection{Learning sparse predictors in the landmarked space}
We use the \emph{Forward Greedy Selection} algorithm presented in \cite{acc-sparse} to extract a sparse predictor in the landmarked space. The algorithm is presented in pseudo code form in Algorithm~\ref{alg-sparse-reg}. The algorithm can be seen as a (modified) form of orthogonal matching pursuit wherein at each step we add a coordinate to the support of the weight vector. The coordinate is added in a greedy manner so as to provide maximum incremental benefit in terms of lowering the loss. Thus the sparsity of the resulting predictor is bounded by the number of steps for which this algorithm is allowed to run. The algorithm requires that it be used with a \emph{smooth} loss function. A loss function $\ell : \R \times \R \rightarrow \R^+$ is said to be $\beta$-smooth if, for all $y,a,b \in \R$, we have 
\[
\ell(a,y) - \ell(b,y) \leq \left.\frac{\partial}{\partial x}\ell(x,y)\right|_{x = b}(a-b) + \frac{\beta(a-b)^2}{2}
\]
Unfortunately, this excludes the $\epsilon$-insensitive loss. However it is possible to run the algorithm with a smooth surrogate whose loss can be transferred to $\epsilon$-insensitive loss. Following \cite{acc-sparse}, we choose the following loss function:
\[
\tilde \ell_\beta(a,b) = \underset{v \in \R}{\inf}\bs{\frac{\beta}{2}v^2 + \eild{a-v,b}}
\]
One can, by a mildly tedious case-by-case analysis, arrive at an explicit form for this loss function
\[
\tilde \ell_\beta(a,b) = \left\{
	\begin{array}{l l}
		0 & \quad \abs{a-b} \leq \epsilon\\
		\frac{\beta}{2}\br{\abs{a-b}- \epsilon}^2 & \quad \epsilon < \abs{a-b} < \epsilon + \frac{1}{\beta}\\
		\abs{a-b} - \epsilon - \frac{1}{2\beta} & \quad \abs{a-b} \geq \epsilon + \frac{1}{\beta}\\
	\end{array}
	\right.
\]
Note that this loss function is convex as well as differentiable (actually $\beta$-smooth) which will be crucial in the following analysis. Moreover, for any $a,b$ we have
\begin{eqnarray}
\label{ell-tilde-ell} 0 \leq \eild{a,b} - \tilde\ell_\beta(a,b) \leq \frac{1}{2\beta}
\end{eqnarray}

{\bf Analysis of Forward Greedy Selection}: We need to setup some notation before we can describe the guarantees given for the predictor learned using the Forward Greedy Selection algorithm. Consider a domain $\X \subset \R^d$ for some $d > 0$ and the class of functions $\F = \bc{\vecx \mapsto \ip{\vecw}{\vecx} : \norm{\vecw}_1 \leq C_W}$. For any distribution $\D$ on $\X$ and any predictor from $\F$, define $\RR_\D(\vecw) := \EE{\vecx \sim \D}{\eild{\ip{\vecw}{\vecx},y(\vecx)}}$ and $\tilde\RR_\D(\vecw) := \EE{\vecx \sim \D}{\tilde \ell_\beta(\ip{\vecw}{\vecx},y(\vecx))}$. Also let $\bar\vecw$ be the minimizer of the following program
\begin{eqnarray}
\label{prog-sparse} \bar\vecw = \underset{\vecw : \norm{\vecw}_1 \leq C_W}{\arg\min}\tilde\RR_\D(\vecw)
\end{eqnarray}

Then \cite[Theorem 2.4]{acc-sparse}, when specialized to our case, guarantees that Algorithm~\ref{alg-sparse-reg}, when executed with $\tilde \ell_\beta(\cdot,\cdot)$ as the loss function for $\beta = \frac{1}{\epsilon_2}$, produces a $k$-sparse predictor $\hat\vecx$, for $k = \left\lceil\frac{8C_W^2}{\epsilon_2^2}\right\rceil$, with $\norm{\hat\vecw}_1 \leq C_W$ such that
\[
\tilde\RR_\D(\hat\vecw) - \tilde\RR_\D(\bar\vecw) \leq \epsilon_2
\]
Thus, if we can show the existence of a good predictor in our space with bounded $L_1$ norm then this would upper bound the loss incurred by the minimizer of Equation~\ref{prog-sparse} and using \cite[Theorem 2.4]{acc-sparse} we would be done. Note that Theorem~\ref{sparse-landmark-guarantee} does indeed give us such a guarantee which allows us to make the following argument: we are guaranteed of the existence of a predictor $\tilde f$ with $L_1$ norm bounded by $B$ that has $\epsilon$-insensitive loss bounded by $(\epsilon_0 + \epsilon_1)$. Thus if we take $C_W = B$ in Equation~\ref{prog-sparse} and use the left inequality of Equation~\ref{ell-tilde-ell}, we get $\tilde \RR_\D(\bar\vecw) \leq \epsilon_0 + \epsilon_1$. Thus we have $\tilde\RR_\D(\hat\vecw) \leq \epsilon_0 + \epsilon_1 + \epsilon_2$. Using Equation~\ref{ell-tilde-ell} (right inequality) with $\beta = \frac{1}{\epsilon_2}$, we get $\RR_\D(\hat\vecw) \leq \epsilon_0 + \epsilon_1 + 3\epsilon_2/2$.

However it is not possible to give utility guarantees with bounded sample complexities using the above analysis, the reason being that Algorithm~\ref{alg-sparse-reg} requires us to calculate, for any given vector $\vecw$, the vector $\nabla_\vecw\tilde\RR(\vecw) = \EE{\vecx \sim \D}{\frac{\partial}{\partial\vecw}\tilde \ell_\beta(\ip{\vecw}{\vecx},y(\vecx))}$ which is infeasible to calculate for a distribution with infinite support since it requires unbounded sample complexities. To remedy we shall, as suggested by \cite{acc-sparse}, take $\D$ not to be the true distribution over the entire domain $\X$, but rather the \emph{empirical distribution} $\D_{\text{emp}} = \frac{1}{n}\sum\limits_{i=1}^n\ind_{\bc{\vecx = \vecx_i}}$ for a given sample of training points $\vecx_1,\ldots,\vecx_n$. Note that the result in \cite{acc-sparse} holds for any distribution which allows us to proceed as before.

Notice however, that we are yet again faced with the challenge of proving an upper bound on the loss incurred by the minimizer of Equation~\ref{prog-sparse}. This we do as follows: the predictor $\tilde f$ defined in Theorem~\ref{sparse-landmark-guarantee} has expected $\epsilon$-insensitive loss over the entire domain bounded by $\epsilon_0 + \epsilon_1$. Hence it will, with probability greater than $\br{1 - \delta}$, have at most $\epsilon_0 + \epsilon_1 + \O{\frac{B}{\sqrt{n}}}$ loss on a random sample of $n$ points by an application of Hoeffding's inequality. Thus we have $\tilde \RR_{\D_{\text{emp}}}(\bar\vecw) \leq \epsilon_0 + \epsilon_1 + \O{\frac{B}{\sqrt{n}}}$ with high probability. 

The main difference in this analysis shall be that the guarantee on $\hat\vecw$ we get will be on its \emph{training loss} rather than its true loss, i.e. we will have $\RR_{\D_{\text{emp}}}(\hat\vecw) \leq \epsilon_0 + \epsilon_1 + \O{\frac{B}{\sqrt{n}}} + \epsilon_2$. However since Algorithm~\ref{alg-sparse-reg} guarantees $\norm{\hat\vecw}_1 \leq C_W = B$, we can still hope to bound its generalization error. More specifically, Lemma~\ref{sparse-risk-bounds}, given below, shows that with probability greater than $\br{1 - \delta}$ over the choice of training points we will have, for all $\vecw \in \R^d$, $\RR_\D(\vecw) - \RR_{\D_{\text{emp}}}(\vecw) \leq \softO{\frac{B}{\sqrt n}}$ where the $\softO{\cdot}$ notation hides certain log factors.

\begin{lem}[Risk bounds for sparse linear predictors \cite{rademacher-averages}]
\label{sparse-risk-bounds}
Consider a real-valued prediction problem $y$ over a domain $\X = \bc{\vecx : \norm{\vecx}_\infty \leq C_X} \subset \R^d$ and a linear learning model $\F : \bc{\vecx \mapsto \ip{\vecw}{\vecx} : \norm{\vecw}_0 \leq k, \norm{\vecw}_1 \leq C_W}$ under some fixed loss function $\ell\br{\cdot,\cdot}$ that is $C_L$-Lipschitz in its second argument. For any $f \in \F$, let $\L_f = \EE{\vecx \sim \D}{\ell(f(\vecx),y(\vecx))}$ and $\hat \L_f^n$ be the empirical loss on a set of $n$ i.i.d. chosen points, then we have, with probability greater than $\br{1 - \delta}$,
\[
\underset{f \in \F}{\sup}\br{\L_f - \hat \L_f^n} \leq 2C_LC_XC_W\sqrt{\frac{2\log(2d)}{n}} + C_LC_XC_W\sqrt{\frac{\log(1/\delta)}{2n}}
\]
\end{lem}

\begin{proof}
The result for non-sparse vectors, that applies here as well, follows in a straightforward manner from \cite[Theorem 1, Example 3.1(2)]{rademacher-averages} and \cite{rademacher-risk-bounds} which we reproduce for completeness. Since the $L_1$ and $L_\infty$ norms are dual to each other, for any $\vecw \in \br{\R^+}^d$ such that $\norm{\vecw}_1 = B$ and any $\mu \in \Delta^d$, where $\Delta^d$ is the probability simplex in $d$ dimensions, the Kullback-divergence function $\text{KL}\br{\left.\frac{\vecw}{B}\right\|\mu}$ is $\frac{1}{B^2}$-strongly convex with respect to the $L_1$ norm. We can remove the positivity constraints on the coordinates of $\vecw$ by using the standard method of introducing additional dimensions that encode negative components of the (signed) weight vector.

Using \cite[Theorem 1]{rademacher-averages}, thus, we can bound the Rademacher complexity of the function class $\F$ as ${\cal R}_n\br{\F} \leq C_XC_W\sqrt{\frac{2\log 2d}{n}}$. Next, using the Lipschitz properties of the loss function, a result from \cite{rademacher-risk-bounds} allows us to bound the excess error by $2C_L{\cal R}_n(\F) + C_LC_XC_W\sqrt{\frac{\log(1/\delta)}{2n}}$. The result then follows.
\end{proof}

Thus, by applying a union bound, with probability at least $\br{1 - 2\delta}$, we will choose a training set such that $\tilde f$, and consequently $\bar\vecw$, has bounded loss on that set as well as the uniform convergence guarantee of Lemma~\ref{sparse-risk-bounds} will hold. Then we can bound the true loss of the predictor returned by Algorithm~\ref{alg-sparse-reg} as
\[
\RR_\D(\hat\vecw) \leq \RR_{\D_{\text{emp}}}(\hat\vecw) + \softO{\frac{B}{\sqrt n}} \leq \epsilon_0 + \epsilon_1 + \epsilon_2 + \softO{\frac{B}{\sqrt n}}
\]
where the first inequality uses the uniform convergence guarantee and the second inequality holds conditional on $\tilde f$ having bounded loss on a given training set. The final guarantee is formally given in Theorem~\ref{good-sim-sparse-guarantee}.

Note that using Lemma~\ref{risk-bounds} here would at best guarantee a decay of $\O{\sqrt{\frac{d}{n}}}$. Transferring $\epsilon$-insensitive loss to absolute loss requires an addition of $\epsilon$. Using all the results given above, we can now give a proof for Theorem~\ref{good-sim-sparse-guarantee} which we restate for convenience.

\begin{thm}[Theorem~\ref{good-sim-sparse-guarantee} restated]
Every similarity function that is $\br{\epsilon_0,B,\tau}$-good for a regression problem with respect to the insensitive loss function $\eild{\cdot,\cdot}$ is $\br{\epsilon_0 + \epsilon}$-useful with respect to absolute loss as well; with the dimensionality of the landmarked space being bounded by $\O{\frac{B^2}{\tau\epsilon_1^2}\log{\frac{1}{\delta}}}$ and the labeled sampled complexity being bounded by $\O{\frac{B^2}{\epsilon_1^2}\log{\frac{B}{\epsilon_1\delta}}}$. Moreover, this utility can be achieved by an $\O{\tau}$-sparse predictor on the landmarked space.
\end{thm}
\begin{proof}
Using Theorem~\ref{sparse-landmark-guarantee}, we first bound the excess loss due to landmarking by $32B\sqrt{\frac{\log(1/\tau\delta)}{d}}$. Next we set up the (dummy) Ivanov regularized regression problem (given in Equation~\ref{prog-sparse}) with the training loss being the objective and regularization parameter $C_W = B$. The training loss incurred by the minimizer of that problem $\vecw_{\text{inter}}$ is, with probability at least $\br{1 - \delta}$, bounded by $\hat\L\br{\vecw_{\text{inter}}} \leq \epsilon_0 + 32B\sqrt{\frac{\log(1/\delta)}{\tau d}} + B\sqrt{\frac{\log(1/\delta)}{n}}$ due to the guarantees of Theorem~\ref{sparse-landmark-guarantee}. Next, we run the Forward Greedy Selection algorithm of \cite{acc-sparse} (specialized to our case in Algorithm~\ref{alg-sparse-reg}) and obtain another predictor $\hat\vecw$ with $L_1$ norm bounded by $B$ that has empirical error at most $\hat\L\br{\hat\vecw} \leq \hat\L\br{\vecw_{\text{inter}}} + \sqrt{\frac{18B^2}{k}}$. Finally, using Lemma~\ref{sparse-risk-bounds}, we bound the true $\epsilon$-insensitive loss incurred by $\hat\vecw$ by $\hat\L\br{\hat\vecw} + 2B\sqrt{\frac{2\log(2d)}{n}} + B\sqrt{\frac{\log(1/\delta)}{2n}}$. Adding $\epsilon$ to convert this loss to absolute loss we get that with probability at most $\br{1 - 3\delta}$, we will output a $k$-sparse predictor in a $d$-dimensional space with absolute regression loss at most
\[
\epsilon_0 + 32B\sqrt{\frac{\log(1/\delta)}{\tau d}} + \sqrt{\frac{18B^2}{k}} + 2B\sqrt{\frac{2\log(2d)}{n}} + 2B\sqrt{\frac{\log(1/\delta)}{2n}} + \epsilon\qedhere
\]
\end{proof}

We note that Forward Greedy Selection gives $\O{\frac{1}{k}}$ error rates, which are much better, if the loss function being used is smooth. This can be achieved by using squared loss $\lsq{a,b} = \br{a-b}^2$ as the surrogate. However we note that assuming goodness of the similarity function in terms of squared loss would impose strictly stronger conditions on the learning problem. This is because $\E{\lsq{a,b}} = \sup\br{a-b}\cdot\E{\abs{a-b}}$ and thus, under boundedness conditions, squared loss is bounded by a constant times the absolute loss but it is not possible to bound absolute loss (or $\epsilon$-insensitive loss) as a constant multiple of the squared loss since there exist distributions such that $\E{\abs{a-b}} = \Om{\frac{1}{\inf\br{\abs{a-b}}}\cdot \E{\lsq{a,b}}}$ and $\frac{1}{\inf\br{\abs{a-b}}}$ can diverge.

Below we prove admissibility results for the sparse learning model.

\subsection{Proof of Theorem~\ref{thm:util_sparse_reg}}
To prove the first part, construct a new weight function $\tilde w(\vecx) = \text{sign}\br{w(\vecx)}\cdot\bar w$. Note that we have $\abs{\tilde w(\vecx)} \leq \bar w \leq B$. Also construct the choice function as follows: for any $\vecx$, let $\Pr{R(\vecx) = 1 | \vecx} = \frac{\abs{w(\vecx)}}{B}$. This gives us $\EE{\vecx \sim \D}{R(\vecx)} = \frac{\bar w}{B}$. Then for any $\vecx$, we have
\begin{eqnarray*}
\EE{\vecx'\sim\D}{\tilde w(\vecx')K(\vecx,\vecx') | R(\vecx')} &=& \EE{\vecx'\sim\D}{\text{sign}\br{w(\vecx)}\bar wK(\vecx,\vecx')\frac{\abs{w(\vecx)}}{B}}/\Prr{\vecx \sim \D}{R(\vecx) = 1}\\
																									 &=& \EE{\vecx'\sim\D}{w(\vecx)K(\vecx,\vecx')\frac{\bar w}{B}}/\br{\frac{\bar w}{B}}\\
																									 &=& \EE{\vecx'\sim\D}{w(\vecx')K(\vecx,\vecx')}
\end{eqnarray*}
Since $f(\vecx) = \EE{\vecx'\sim\D}{w(\vecx')K(\vecx,\vecx')}$ has small $\epsilon$-insensitive loss by $\br{\epsilon_0,B}$-goodness of $K$, we have our result.
To prove the second part, construct a new weight function $\tilde w(\vecx) = \frac{w(\vecx)}{\tau}\Pr{R(\vecx) = 1 | \vecx}$. Note that we have $\abs{\tilde w(\vecx)} \leq \frac{B}{\tau}$. Then for any $\vecx$, we have
\begin{eqnarray*}
\EE{\vecx'\sim\D}{\tilde w(\vecx')K(\vecx,\vecx')} &=& \EE{\vecx'\sim\D}{\frac{w(\vecx')}{\tau}R(\vecx')K(\vecx,\vecx')}\\
																									 &=& \EE{\vecx'\sim\D}{\frac{w(\vecx')}{\tau}K(\vecx,\vecx') | R(\vecx')}\Prr{\vecx'\sim\D}{R(\vecx') = 1}\\
																									 &=& \EE{\vecx'\sim\D}{w(\vecx')K(\vecx,\vecx')| R(\vecx')}
\end{eqnarray*}

Since $f(\vecx) = \EE{\vecx'\sim\D}{w(\vecx')K(\vecx,\vecx') | R(\vecx')}$ has small $\epsilon$-insensitive loss by $\br{\epsilon_0,B,\tau}$-goodness of $K$, we have our result.

Using the above result we get out admissibility guarantee.
\begin{cor}
Every PSD kernel that is $\br{\epsilon_0,\gamma}$-good for a regression problem is, for any $\epsilon_1 > 0$, $\br{\epsilon_0 + \epsilon_1, \O{\frac{1}{\epsilon_1 \gamma^2}},1}$-good as a similarity function as well.
\end{cor}
The above result is rather weak with respect to the sparsity parameter $\tau$ since we have made no assumptions on the distribution of the dual variables $\alpha_i,\alpha_i^\ast$ in the proof of Theorem~\ref{reg-psd-sim-good} which is why we are forced to use the (weak) inequality $\frac{\bar w}{B} \leq 1$. Any stronger assumptions on the kernel goodness shall also strengthen this admissibility result.

\section{Ordinal Regression}
In this section we give missing utility and admissibility proofs for the similarity-based learning model for ordinal regression. But before we present the analysis of our model, we give below, an analysis of algorithms that choose to directly reduce the ordinal regression problem to real-valued regression. The analysis will serve as motivation that will help us define our goodness criteria.

\subsection{Reductions to real valued regression}
One of the simplest learning algorithms for the problem of ordinal regression involves a reduction to real-valued regression \cite{ordreg-guarantee, ordreg-svm} where we modify our goal to that of learning a real valued function $f$ which we then threshold using a set of thresholds $\bc{b_i}_{i=1}^{r}$ with $b_1 = -\infty$ to get discrete labels as shown below
\[
y_f(\vecx) = \underset{i \in \bs{r}}{\mathop{\arg\max}}\bc{b_i : f(\vecx) \geq b_i}
\]
These thresholds may themselves be learned or fixed apriori. A simple choice for these thresholds is $b_i = i-1$ for $i > 1$. It is easy to show (using a result in \cite{ordreg-guarantee}) that for the fixed thresholds specified above, we have for all $f : \X \rightarrow \R$,
\begin{eqnarray*}
\lord{y_f(\vecx),y(\vecx)} &\leq& \min\bc{2\abs{f(\vecx) - y(\vecx)}, \abs{f(\vecx) - y(\vecx)} + \frac{1}{2}}\\
&\leq& \min\bc{2\eild{f(\vecx) - y(\vecx)} + 2 \epsilon, \eild{f(\vecx) - y(\vecx)} + \epsilon + \frac{1}{2}}
\end{eqnarray*}
where in the last step we use the fact that $\abs{x} - \epsilon \leq \eild{x} \leq \abs{x}$.

It is tempting to use this reduction along with guarantees given for real-valued regression to directly give generalization bounds for ordinal regression. To pursue this further, we need a notion of a good similarity function which we give below:

\begin{defn}
A similarity function $K$ is said to be $\br{\epsilon_0,B}$-good for an ordinal regression problem $y : \X \rightarrow \bs{r}$ if for some bounded weight function $w : \X \rightarrow \bs{-B,B}$, the following predictor, when subjected to fixed thresholds, has expected ordinal regression error at most $\epsilon_0$
\[
f : \vecx \mapsto \EE{\vecx' \sim \D}{w(\vecx')K(\vecx,\vecx')}
\]
i.e. $\EE{\vecx \sim \D}{\abs{y_f(\vecx) - y(\vecx)}} < \epsilon_0$.
\end{defn}

From the definition of the thresholding scheme used to define $y_f$ from $f$, it is clear that $\abs{f(\vecx) - y(\vecx)} \leq \abs{y_f(\vecx) - y(\vecx)} + \frac{1}{2}$. Since we have $\eild{x} \leq \abs{x}$ for any $\epsilon \geq 0$, we have $\eild{f(\vecx) - y(\vecx)} \leq \abs{y(\vecx) - y_f(\vecx)} + \frac{1}{2}$ and thus we have $\EE{\vecx \sim \D}{\eild{f(\vecx), y(\vecx)}} < \epsilon_0 + \frac{1}{2}$.

Thus, starting with goodness guarantee of the similarity function with respect to ordinal regression, we obtain a guarantee of the goodness of the similarity function $K$ with respect to real-valued regression that satisfies the requirements of Theorem~\ref{good-sim-guarantee}. Thus we have the existence of a linear predictor over a low dimensional space with $\epsilon$-insensitive error at most $\epsilon_0 + \frac{1}{2} + \epsilon_1$. We can now argue (using results from \cite{ordreg-guarantee}) that this real-valued predictor, when subjected to the fixed thresholds, would yield a predictor with ordinal regression error at most
\[
\min\bc{2\br{\epsilon_0 + \frac{1}{2} + \epsilon_1} + 2\epsilon, \br{\epsilon_0 + \frac{1}{2} + \epsilon_1} + \epsilon + \frac{1}{2}} = 1 + \epsilon_0 + \epsilon_1 + \epsilon.
\]
However, this is rather disappointing since this implies that the resulting predictor would, on an average, give out labels that are at least one step away from the true label. This forms the intuition behind introducing (soft) margins in the goodness formulation that gives us Definition~\ref{def:oreg-good}. Below we give proofs for utility and admissibility guarantees for our model for similarity-based ordinal regression.

\subsection{Proof of Theorem~\ref{good-sim-guarantee-ord}}
We use Lemma~\ref{landmark-guarantee} to construct a landmarked space with a linear predictor $\tilde f : \vecx \mapsto \ip{\vecw}{\vecx}$ such that $\EE{\vecx \sim \D}{\abs{\tilde f\br{\Psi\br{\vecx}} - f(\vecx)}} \leq 2\epsilon_1$. As before, we have $\norm{\vecw}_2 \leq B$ and $\underset{\vecx \in \X}{\sup}\bc{\norm{\Psi(\vecx)}} \leq 1$. In the following, we shall first show bounds on the mislabeling error i.e $\Prr{\vecx \sim \D}{\hat y(\vecx) \neq y(\vecx)}$. Next, we shall convert these bounds into ordinal regression loss by introducing a \emph{spacing} parameter into the model.

Since the $\gamma$-margin loss function is $1$-Lipschitz, we get
\begin{eqnarray*}
\gmld{\tilde f(\Psi(\vecx)) - b_{y(\vecx)}} &\leq& \gmld{f(\vecx) - b_{y(\vecx)}} + 2\epsilon_1\\
\gmld{b_{y(\vecx) + 1} - \tilde f(\Psi(\vecx))} &\leq& \gmld{b_{y(\vecx) + 1} - f(\vecx)} + 2\epsilon_1
\end{eqnarray*}
Which gives us, upon taking expectations on both sides,
\[
\EE{\vecx \sim \D}{\gmld{\tilde f(\Psi(\vecx)) - b_{y(\vecx)}} + \gmld{b_{y(\vecx) + 1} - \tilde f(\Psi(\vecx))}} \leq \epsilon_0 + 4\epsilon_1
\]
Lemma~\ref{landmark-guarantee} guarantees the excess loss due to landmarking to be at most $64B\sqrt{\frac{\log(1/\delta)}{d}}$. Moreover, since the $\gamma$-margin loss is $1$-Lipschitz, Lemma~\ref{risk-bounds} allows us to bound excess loss due to training by $3B\sqrt{\frac{\log(1/\delta)}{n}}$ so that the learned predictor has $\gamma$-margin loss at most $\epsilon_0 + \epsilon_1$ for any $\epsilon_1$ given large enough $d$ and $n$. Now, from the definition of the $\gamma$-margin loss it is clear that if the loss is greater than $\gamma$ then it indicates a mislabeling. Hence, the mislabeling error is bounded by $\frac{\epsilon_0 + \epsilon_1}{\gamma}$.

This may be unsatisfactory if $\gamma \ll 1$ - to remedy such situations we show that we can bound the $1$-margin loss directly.
Starting from $\EE{\vecx \sim \D}{\abs{\tilde f(\Psi(\vecx)) - f(\vecx)}} < 2 \epsilon_1$, we can also deduce
\[
\EE{\vecx \sim \D}{\pos{1 - \tilde f(\Psi(\vecx)) + b_{y(\vecx)}} + \pos{1 - b_{y(\vecx) + 1} + \tilde f(\Psi(\vecx))}} \leq \epsilon_0 + 4 \epsilon_1
\]
We can bound the excess training error for this loss function as well. Since the $1$-margin loss directly bounds the mislabeling error, combining the two arguments we get the second part of the claim.

However, the margin losses themselves do not present any bound on the ordinal regression error. This is because, if the thresholds are closely spaced together, then even an instance of gross ordinal regression loss could correspond to very small margin loss. To remedy this, we introduce a \emph{spacing} parameter into the model. We say that a set of thresholds is $\Delta$-spaced if $\underset{i \in \bs{r}}{\min}\bc{\abs{b_i - b_{i+1}}} \geq \Delta$. Such a condition can easily be incorporated into the model of \cite{ordreg-guarantee} as a constraint in the optimization formulation.

Suppose that a given instance has ordinal regression error $\lord{\hat y(\vecx),y(\vecx)} = k$. This can happen if the point was given a label $k$ labels below (or above) its correct label. Also suppose that the $\gamma$-margin error in this case is $\gmld{\hat y(\vecx) - y(\vecx)} = h$. Without loss of generality, assume that the point $\vecx$ of label $k+1$ was given the label $1$ giving an ordinal regression loss of $l_{\text{ord}} = k$ (a similar analysis would hold if the point of label $1$ were to be given a label $k+1$ by symmetry of the margin loss formulation with respect to left and right thresholds). In this case the value of the underlying regression function must lie between $b_1$ and $b_2$ and thus, the margin loss $h$ satisfies $h \geq b_{k+1} + \gamma - b_2 = \gamma + \sum\limits_{i=2}^{k}\br{b_{i+1} - b_i} \geq \gamma + \br{k-1}\Delta$. Thus, if the margin loss is at most $h$, the ordinal regression error must satisfy $\lord{\hat y(\vecx), y(\vecx)} \leq \frac{\gmld{\hat y(\vecx) - b_{y(\vecx)}} + \gmld{b_{y(\vecx) + 1} - \hat y(\vecx)} - \gamma}{\Delta} + 1$. Let $\psi_\Delta(x) = \frac{x + \Delta - 1}{\Delta}$. Using the bounds on the $\gamma$-margin and $1$-margin losses given above, we get the first part of the claim.

In particular, a constraint of $\Delta = 1$ put into an optimization framework ensures that the bounds on mislabeling loss and ordinal regression loss match since $\psi_1(x) = x$ for all $x$. In general, the cases where the above framework yields a non-trivial bound for the mislabeling error rate, i.e. $\ell_{01} < 1$ (which can always be ensured if $\epsilon_0 < 1$ by taking large enough $d$ and $n$), also correspond to those where the ordinal regression error rate is also bounded above by $1$ since $\underset{x \in \bs{0,1},\Delta > 0}{\sup}\br{\psi_\Delta\br{x}} = 1$.

\subsection{Admissibility Guarantees}
\label{app:oreg_adm}
We begin by giving the kernel goodness criterion which we adapt from existing literature on large margin approaches to ordinal regression. More specifically we use the framework described in \cite{ordreg-svm} for which generalization guarantees are given in \cite{ordreg-guarantee}.
\begin{defn}
Call a PSD kernel $K$ $\br{\epsilon_0,\gamma}$-good for an ordinal regression problem $y : \X \rightarrow \bs{r}$ if there exists $\vecW^\ast \in \H_K$, $\norm{\vecW^\ast} = 1$ and a fixed set of thresholds $\bc{b_i}_{i=1}^{r}$ such that
\[
\EE{\vecx \sim \D}{\pos{b_{y(\vecx)} + 1 - \frac{\ip{\vecW^\ast}{\Phi_K(\vecx)}}{\gamma}} + \pos{\frac{\ip{\vecW^\ast}{\Phi_K(\vecx)}}{\gamma} - b_{y(\vecx) + 1} + 1}} < \epsilon_0
\]
\end{defn}
The above definition exactly corresponds to the EXC formulation put forward by \cite{ordreg-guarantee} except for the fact that during actual optimization, a strict ordering on the thresholds is imposed explicitly. \cite{ordreg-guarantee} present yet another model called IMC which does not impose any explicit orderings, rather the ordering emerges out of the minimization process itself. Our model can be easily extended to the IMC formulation as well.

\begin{thm}[Theorem~\ref{thm:adm_oreg} restated]
Every PSD kernel that is $\br{\epsilon_0,\gamma}$-good for an ordinal regression problem is also $\br{\gamma_1\epsilon_0 + \epsilon_1, \O{\frac{\gamma_1^2}{\epsilon_1 \gamma^2}}}$-good as a similarity function with respect to the $\gamma_1$-margin loss for any $\gamma_1, \epsilon_1 > 0$. Moreover, for any $\epsilon_1 < \gamma_1/2$, there exists an ordinal regression instance and a corresponding kernel that is $\br{0,\gamma}$-good for the ordinal regression problem but only $\br{\epsilon_1,B}$-good as a similarity function with respect to the $\gamma_1$-margin loss function for $B = \Om{\frac{\gamma_1^2}{\epsilon_1\gamma^2}}$.
\end{thm}
\begin{proof}
We prove the two parts of the result separately.

{\bf Part 1: Admissibility}: As before, using Lemma~\ref{admissible-weights} it is possible to obtain a vector $\vecW' = \sum\limits_{i=1}^n(\alpha_i - \alpha_i^\ast)\Phi_K(\vecx_i) \in \H_K$ such that $0 \leq \alpha_i,\alpha_i^\ast \leq p_iC$ (by applying the KKT conditions) and the following holds:
\begin{eqnarray}
\label{eq1} \EE{\vecx \sim \D}{\pos{b_{y(\vecx)} + 1 - \ip{\vecW'}{\Phi_K(\vecx)}} + \pos{\ip{\vecW'}{\Phi_K(\vecx)} - b_{y(\vecx) + 1} + 1}} < \frac{1}{2C\gamma^2} + \epsilon_0
\end{eqnarray}
This allows us to construct a weight function $w_i = \frac{\alpha_i - \alpha_i^\ast}{p_i}$ such that $\abs{w_i} \leq 2C$ (since we do not have any guarantee that $\alpha_i\alpha^\ast_i = 0$) and $\EE{\vecx' \sim \D}{w(\vecx')K(\vecx,\vecx')} = \ip{\vecW'}{\Phi_K(\vecx)}$ for all $\vecx \in \X$. Denoting $f(\vecx) := \EE{\vecx' \sim \D}{w(\vecx')K(\vecx,\vecx')}$ for convenience gives us
\begin{eqnarray*}
\EE{\vecx \sim \D}{\gml{1}{f(\vecx) - b_{y(\vecx)}} + \gml{1}{b_{y(\vecx) + 1} - f(\vecx)}} &=& \EE{\vecx \sim \D}{\pos{1 - f(\vecx) + b_{y(\vecx)}} + \pos{1 - b_{y(\vecx) + 1} + f(\vecx)}}\\
&\leq& \frac{1}{2C\gamma^2} + \epsilon_0
\end{eqnarray*}
where in the first step we used $\gml{1}{x} = \pos{1 - x}$. Now use the fact $\gml{1}{x} = \frac{1}{\gamma}\gmld{\gamma x}$ to get the following:
\[
\EE{\vecx \sim \D}{\gml{\gamma_1}{\gamma_1 f(\vecx) - \gamma_1 b_{y(\vecx)}} + \gml{\gamma_1}{\gamma_1 b_{y(\vecx) + 1} - \gamma_1 f(\vecx)}} \leq \frac{\gamma_1}{2C\gamma^2} + \gamma_1\epsilon_0
\]
Note that it is not possible to perform the analysis on the loss function $\gml{\gamma}{\cdot}$ directly since using it requires us to scale the threshold values by a factor of $\gamma_1$ that makes the result in Equation~\ref{eq1} unusable. Hence we first perform the analysis for $\gml{1}{\cdot}$, utilize Equation~\ref{eq1} and then interpret the resulting inequality in terms of $\gml{\gamma_1}{\cdot}$.

Setting $2C = \frac{\gamma_1}{\epsilon_1\gamma^2}$, using $w'(\vecx) = \gamma_1 w(\vecx)$ as weights, using $b'_j = \gamma_1 b_j$ as the thresholds and noting that the new bound on the weights is $\abs{w'_i} \leq 2C\gamma_1$ gives us the result. As before, using variational optimization techniques, this result can be extended to non-discrete distributions as well.
\end{proof}

In particular, setting $\gamma_1 = \gamma$ gives us that any PSD kernel that is $\br{\epsilon_0,\gamma}$-good for an ordinal regression problem is also  $\br{\gamma\epsilon_0 + \epsilon_1, \frac{1}{\epsilon_1}}$-good as a similarity function with respect to the $\gamma$-margin loss.

{\bf Part 2: Tightness}: We adapt our running example (used for proving the lower bound for real regression) for the case of ordinal regression as well. Consider the points with value $-1$ as having label $1$ and those having value $+1$ as having label $2$. Clearly, $w = \br{1,0,0}$ along with the thresholds $b_1 = -\infty$ and $b_2 = 0$ establishes the native inner product as a $\br{0,\gamma}$-good PSD kernel.

Now consider the heavy points yet again and some weight function and threshold $b_2$ ($b_1$ is always fixed at $-\infty$) that is supposed to demonstrate the goodness of the inner product kernel as a similarity function. Clearly we have
\begin{eqnarray*}
\EE{\vecx \sim \D}{\gml{\gamma_1}{f(\vecx) - b_{y(\vecx)}} + \gml{\gamma_1}{b_{y(\vecx) + 1} - f(\vecx)}} &\geq& \br{\frac{1}{2} - \epsilon}\br{\gml{\gamma_1}{f(\vecx_1) - b_2} + \gml{\gamma_1}{b_2 - f(\vecx_4)}}\\
 &=& \br{\frac{1}{2} - \epsilon}\br{\pos{\gamma_1 - f(\vecx_1) + b_2} + \pos{\gamma_1 - b_2 + f(\vecx_4)}}\\
 &\geq& \br{\frac{1}{2} - \epsilon}\br{2\gamma_1 - f(\vecx_1) + f(\vecx_4)}\\
 &=& \br{\frac{1}{2} - \epsilon}\br{2\gamma_1 - \br{\frac{1}{2} - \epsilon}\br{1-b}\br{w_4 - w_1}}\\
 &=& \br{\frac{1}{2} - \epsilon}\br{2\gamma_1 - \br{\frac{1}{2} - \epsilon}\br{4\gamma^2}\br{w_4 - w_1}}
\end{eqnarray*}
where in the third step we have used the fact that $\pos{a} + \pos{b} \geq a + b$. Thus, in order to have expected error at most $\epsilon_1$, we must have
\[
w_4 - w_1 \geq \frac{1}{4\gamma^2}\br{2\gamma_1 - \frac{\epsilon_1}{\frac{1}{2} - \epsilon}}\frac{1}{\frac{1}{2} - \epsilon} = \frac{\gamma_1^2}{4 \epsilon_1\gamma^2}
\]
by setting $\epsilon = \frac{1}{2} - \frac{\epsilon_1}{\gamma_1}$ which then proves the result after applying an averaging argument.

\section{Ranking}
\label{app:rank}
The problem of ranking stems from the need to sort a set of items based on their relevance. In the model considered here, each ranking instance is composed of $m$ documents (pages) $\br{p_1,\ldots,p_m}$ from some universe $\P$ along with their relevance to some particular query $q \in \Q$ that are given as relevance scores from some set $\RR \subset \R$. Thus we have $\X = \Q \times \P^m$ with each instance $\vecx \in \X$ being provided with a relevance vector $r(\vecx) = \RR^m$. Let the $i^{th}$ query-document pair of a ranking instance $\vecx$ be denoted by $\vecz_i \in \Q \times \P$. For any $\vecz = (p,q) \in \P \times \Q$, let $r(\vecz) \in \R$ denote the true relevance of document $p$ to query $q$.

For any relevance vector $\vecr \in \RR^m$, let $\bar\vecr$ be the vector with elements of $\vecr$ sorted in descending order and $\pi_\vecr$ be the permutation that this sorting induces. For any permutation $\pi$, $\pi(i)$ shall denote the index given to the index $i$ under $\pi$. Although the desired output of a ranking problem is a permutation, we shall follow the standard simplification \cite{ndcg} of requiring the output to be yet another relevance vector $\vecs$ with the permutation $\pi_\vecs$ being considered as the actual output. This converts the ranking problem into a vector-valued regression problem.

We will take the true loss function $\ellac{\cdot}{\cdot}$ to be the popular NDCG loss function \cite{ndcg-def} defined below
\[
\lndcg{\vecs,\vecr} = -\frac{1}{\norm{G(\vecr)}_D}\sum\limits_{i=1}^m \frac{G(\vecr(i))}{F(\pi_\vecs(i))}
\]
where $\displaystyle \norm{\vecr}_D = \underset{\pi \in S_m}{\max}\sum\limits_{i=1}^m \frac{\vecr(i)}{F(\pi(i))}$, $G(r) = 2^r - 1$ is the growth function and $F(t) = \log(1 + t)$ is the decay function.

For the surrogate loss functions $\ell_K$ and $\ell_S$, we shall use the squared loss function $\lsq{\vecs,\vecr} = \norm{\vecs - \vecr}_2^2$. We shall overload notation to use $\lsq{\cdot,\cdot}$ upon reals as well. For any vector $\vecr \in \RR^m$, let $\displaystyle \eta(\vecr) := \frac{G(\vecr)}{\norm{G(\vecr)}_D}$ and let $\vecr_i$ denote its $i^{th}$ coordinate.

Due to the decomposable nature of the surrogate loss function, we shall require kernels and similarity functions to act over query-document pairs i.e. $K : \br{\P \times \Q} \times \br{\P \times \Q} \rightarrow \R$. This also coincides with a common feature extraction methodology (see for example \cite{ndcg, rank-stability}) where every query-document pair is processed to yield a feature vector. Consequently, all our goodness definitions shall loosely correspond to the ability of a kernel/similarity to accurately predict the true relevance scores for a given query-document pair. We shall assume ranking instances to be generated by the sampling of a query $q \sim \D_\Q$ followed by $m$ independent samples of documents from the (conditional) distribution $\D_{\P | q}$. The distribution over ranking instances is then a product distribution $\D = \D_\X = \D_\Q \times \underbrace{\D_{\P | q} \times \D_{\P | q} \times \ldots \times \D_{\P | q}}_\text{$m$ times}$. A key consequence of this generative mechanism is that the $i^{th}$ query-document pair of a random ranking instance, for any fixed $i$, is a random query-document instance selected from the distribution $\mu := \D_\Q \times \D_{\P | q}$.


\begin{defn}
\label{sim-good-rank}
A similarity function $K$ is said to be $\br{\epsilon_0,B}$-good for a ranking problem $y : \X \rightarrow S_m$ if for some bounded weight function $w : \P \times \Q \rightarrow \bs{-B,B}$, for any ranking instance $\vecx = \br{q, p_1, p_2, \ldots, p_m}$, if we define $f : \X \rightarrow \R^m$ as
\[
f_i := \EE{\vecz \sim \mu}{w(\vecz)K(\vecz_i,\vecz)}
\]
where $\vecz_i = (p_i,q)$, then we have $\EE{\vecx \sim \D}{\lsq{f(\vecx),\eta(r(\vecz))}} < \epsilon_0$.
\end{defn}

\begin{defn}
\label{kernel-good-rank}
A PSD kernel $K$ is said to be $\br{\epsilon_0,\gamma}$-good for a ranking problem $y : \X \rightarrow S_m$ if there exists $\vecW^\ast \in \H_K$, $\norm{\vecW^\ast} = 1$ such that if for any ranking instance $\vecx = \br{q, p_1, p_2, \ldots, p_m}$, if, for any $\vecW \in \H_K$, when we define $f\br{\ \cdot\ ;\vecW} : \X \rightarrow \R^m$ as
\[
f_i(\vecx;\vecW) = \frac{\ip{\vecW}{\Phi_K(\vecz_i)}}{\gamma}
\]
where $f_i$ is the $i^{th}$ coordinate of the output of $f$ and $\vecz_i = (p_i,q)$, then we have $\EE{\vecx \sim \D}{\lsq{f(\vecx;\vecW^\ast),\eta(r(\vecz))}} < \epsilon_0$.
\end{defn}

The choice of this surrogate is motivated by consistency considerations. We would ideally like a minimizer of the surrogate loss to have bounded actual loss as well. Using results from \cite{ndcg}, it can be shown that the above defined surrogate is not only consistent, but that excess loss in terms of this surrogate can be transferred to excess loss in terms of $\lndcg{\cdot,\cdot}$, a very desirable property. Although \cite{ndcg} shows this to be true for a whole family of surrogates, we chose $\lsq{\cdot,\cdot}$ for its simplicity. All our utility arguments carry forward to other surrogates defined in \cite{ndcg} with minimal changes.

We move on to prove utility guarantees for the given similarity learning model.

\begin{thm}
\label{good-sim-guarantee-rank}
Every similarity function that is $\br{\epsilon_0,B}$-good for a ranking problem for $m$-documents with respect to squared loss is $\O{\sqrt{\frac{m}{\log m}}\cdot\sqrt{\epsilon_0}}$-useful with respect to NDCG loss.
\end{thm}
\begin{proof}
As before, we use Lemma~\ref{landmark-guarantee} to construct a landmarked space with a linear predictor $\tilde f : \vecx \mapsto \ip{\vecw}{\vecx}$ such that $\EE{\vecz \sim \mu}{\abs{\tilde f\br{\Psi\br{\vecz}} - f(\vecz)}} \leq 2\epsilon_1$. We have $\norm{\vecw}_2 \leq B$ and $\underset{\vecx \in \X}{\sup}\bc{\norm{\Psi(\vecx)}} \leq 1$. Now lets overload notation to denote by $\Psi(\vecx)$ the concatenation of the images of the $m$ document-query pairs in $\vecx$ under $\Psi(\cdot)$ and by $\tilde f(\Psi(\vecx))$, the $m$-dimensional vector obtained by applying $\tilde f$ to each of the $m$ components of $\Psi(\vecx)$.

Since the squared loss function is $2B$-Lipschitz in its first argument in the region of interest, we get
\begin{eqnarray*}
\EE{\vecx \sim \D}{\lsq{\tilde f(\Psi(\vecx)),\eta(r(\vecx))}} &=& \EE{\vecx \sim \D}{\sum\limits_{i=1}^m{\lsq{\tilde f(\Psi(\vecz_i)), \eta(r(\vecx))_i}}}\\
&=& \sum\limits_{i=1}^m{\EE{\vecx \sim \D}{\lsq{\tilde f(\Psi(\vecz_i)), \eta(r(\vecx))_i}}}\\
&=& \sum\limits_{i=1}^m\EE{\vecx \sim \D}{\lsq{f(\vecz_i), \eta(r(\vecx))_i}} + \\
&& \sum\limits_{i=1}^m\EE{\vecx \sim \D}{\lsq{\tilde f(\Psi(\vecz_i)), \eta(r(\vecx))_i} - \lsq{f(\vecz_i), \eta(r(\vecx))_i}}\\
&\leq& \sum\limits_{i=1}^m\EE{\vecx \sim \D}{\lsq{f(\vecz_i), \eta(r(\vecx))_i}} + 2B\sum\limits_{i=1}^m\EE{\vecx \sim \D}{\abs{\tilde f(\Psi(\vecz_i)) -f(\vecz_i)}}\\
&=& \sum\limits_{i=1}^m\EE{\vecx \sim \D}{\lsq{f(\vecz_i), \eta(r(\vecx))_i}} + 2B\sum\limits_{i=1}^m\EE{\vecz \sim \mu}{\abs{\tilde f(\Psi(\vecz)) -f(\vecz)}}\\
&\leq& \sum\limits_{i=1}^m\EE{\vecx \sim \D}{\lsq{f(\vecz_i), \eta(r(\vecx))_i}} + 4Bm \epsilon_1\\
&=& \EE{\vecx \sim \D}{\sum\limits_{i=1}^m\lsq{f(\vecz_i), \eta(r(\vecx))_i}} + 4Bm \epsilon_1\\
&=& \EE{\vecx \sim \D}{\lsq{f(\vecx), \eta(r(\vecx))}} + 4Bm \epsilon_1\\
&\leq& \epsilon_0 + 4Bm \epsilon_1
\end{eqnarray*}

where $\vecx = \br{q,p_1,\dots,p_m}$ and $\vecz_i = (p_i,q)$. In the first and the last but one step we have used decomposability of the squared loss, in the fourth step we have used Lipschitz properties of the squared loss, in the fifth step we have used properties of the generative mechanism assumed for ranking instances, in the sixth step we have used the guarantee given by Lemma~\ref{landmark-guarantee}. Throughout we have repeatedly used linearity of expectation. This bounds the excess error due to landmarking to $d$ dimensions by $64B^2m^2\sqrt{\frac{\log(1/\delta)}{d}}$ using Lemma~\ref{landmark-guarantee}. Similarly, Lemma~\ref{risk-bounds} also allows us to bound the excess error due to training by $3B^2\sqrt{\frac{\log(1/\delta)}{n}}$ which puts our total squared loss at $\epsilon_0 + \epsilon_1$ for large enough $d$ and $n$.

We now invoke \cite[Theorem 10]{ndcg} that states that if the surrogate loss function $\ell(\cdot,\cdot)$ being used is a Bregman divergence generated by a function that is $C_S$-strongly convex with respect to some norm $\norm{\cdot}$ then we can bound $\lndcg{\vecs,\vecr} \leq \frac{C_F}{\sqrt{C_S}}\cdot\sqrt{\ell\br{\vecs,\vecr}}$ where $C_F = 2\norm{\br{\frac{1}{F(1)},\ldots,\frac{1}{F(m)}}^\top}_\ast$, $F$ is the decay function used in the definition of NDCG and $\norm{\cdot}_\ast$ is the dual norm of $\norm{\cdot}$. Note that we are using the ``noiseless'' version of the result where $r(\vecx)$ is a deterministic function of $\vecx$.

In our case the squared loss is $2$-strongly convex with respect to the $L_2$ norm which is its own dual. Hence $C_S = 2$ and $C_F = \O{\sqrt{\frac{m}{\log m}}}$, if $\hat f : \vecx \mapsto \ip{\hat \vecw}{\Psi(\vecx)}$ is our final output, we get, for some constant $C$,
\[
\EE{\vecx \sim \D}{\lndcg{\hat f(\vecx), r(\vecx)}} \leq C\sqrt{\frac{m}{\log m}}\cdot\sqrt{\epsilon_0 + 4Bm\epsilon_1} \leq C\sqrt{\frac{m}{\log m}}\cdot\sqrt{\epsilon_0} + C\frac{2m}{\sqrt{\log m}}\cdot\sqrt{B\epsilon_1}
\]
which proves the claim. This affects the bounds given by Lemmata~\ref{landmark-guarantee} and \ref{risk-bounds} since the dependence of the excess error on $d$ and $n$ will now be in terms of the inverse of their fourth roots instead of inverse of the square roots as was the case in regression and ordinal regression.
\end{proof}

We note that the (rather heavy) dependence of the final utility guarantee (that is $\O{\sqrt{m\epsilon_0}}$) on $m$ is because the decay function $F(t) = \log(1+t)$ chosen here (which seems to be a standard in literature but with little theoretical justification) is a very slowly growing function (it might sound a bit incongruous to have an increasing function as our \emph{decay} function - however since this function appears in the denominator in the definition of NDCG, it effectively induces a decay). Using decay functions that grow super-linearly (or rather those that induce super-linear decays), we can ensure $\O{\sqrt{\epsilon_0}}$-usefulness since in those cases, $C_F = \O{1}$.

We next prove admissibility bounds for the ranking problem. The learning setting as well as the proof is different for ranking (due to presence of multiple entities in a single ranking instance), hence we shall provide all the arguments for completeness. 
\begin{thm}
Every PSD kernel that is $\br{\epsilon_0,\gamma}$-good for a ranking problem is also $\br{\epsilon_0 + \epsilon_1, \O{\frac{m\sqrt{m}}{	\epsilon_1\sqrt{\epsilon_1}\gamma^3}}}$-good as a similarity function for any $\epsilon_1 > 0$.
\end{thm}
\begin{proof}

For notational convenience, we shall assume that the RKHS $\H_K$ is finite dimensional so that we can talks in terms of finite dimensional matrices and vectors. As before, let $f(\vecz;\vecW) = \ip{\vecW}{\Phi_K(\vecz)}$ and let $\vecW'$ be the minimizer of the following program.
\begin{eqnarray*}
\underset{\vecW \in \H_K} \min	&& \frac{1}{2}\norm{\vecW}_{\H_K}^2 + C\EE{\vecx \sim \D}{\lsq{f(\vecx;\vecW),\eta(r(\vecx))}}\\
\equiv \qquad \underset{\vecW \in \H_K} \min	&& \frac{1}{2}\norm{\vecW}_{\H_K}^2 + C\EE{\vecx \sim \D}{\sum\limits_{i=1}^m{\lsq{f(\vecz_i;\vecW),\eta(r(\vecx))_i}}}\\
\equiv \qquad \underset{\vecW \in \H_K} \min	&& \frac{1}{2}\norm{\vecW}_{\H_K}^2 + C\sum\limits_{i=1}^m\EE{\vecx \sim \D}{\lsq{f(\vecz_i;\vecW),\eta(r(\vecx))_i}}\\
\equiv \qquad \underset{\vecW \in \H_K} \min	&& \frac{1}{2}\norm{\vecW}_{\H_K}^2 + mC\EE{\vecz \sim \mu}{\lsq{f(\vecz;\vecW),\tilde r(\vecz)}} + C_\D
\end{eqnarray*}
where for any $\vecz \in \Q \times \P$, $\tilde r(\vecz)$ gives us the expected normalized relevance of this document-query pair across ranking instances and $C_\D$ is some constant independent of $\vecW$ and dependent solely on the underlying distributions.
Using the goodness of the kernel $K$ and the argument given in the proof of Lemma~\ref{admissible-weights}, it is possible to show that the vector $\vecW'$ has squared loss at most $\frac{1}{2C\gamma^2} + \epsilon_0$. Hence the only task remaining is to show that their exists a bounded weight function $w$ such that for all $\vecz \in \P \times \Q$, we have $f(\vecz;\vecW) = \ip{\vecW'}{\Phi_K(\vecz)} = \EE{\vecz' \sim \mu}{w(\vecz)K(\vecz,\vecz')}$ which will prove the claim.

To do so we assume that the (finite) set of document-query pairs is $\br{\vecz_1,\ldots,\vecz_k}$ with $\vecz_i$ having probability $\mu_i$ and relevance $r_i = \tilde r(\vecz_i)$. Then the above program can equivalently be written as
\begin{eqnarray*}
\underset{\vecW \in \H_K} \min	&& \frac{1}{2}\norm{\vecW}_{\H_K}^2 + mC\sum\limits_{i=1}^k \mu_i\lsq{\ip{\vecW}{\Phi_K(\vecz_i)}, r_i}\\
\equiv \qquad \underset{\vecW \in \H_K} \min	&& \frac{1}{2}\norm{\vecW}_{\H_K}^2 + mC\norm{\sqrt{P}X^\top\vecW - \sqrt{P}\vecr}_2^2\\
\equiv \qquad \underset{\vecW \in \H_K} \min	&& \frac{1}{2}\norm{\vecW}_{\H_K}^2 + mC\norm{\tilde X^\top\vecW - \tilde\vecr}_2^2\\
\equiv \qquad \underset{\alpha \in \R^{mn}} \min	&& \frac{1}{2}\norm{X\alpha}_{\H_K}^2 + mC\norm{\tilde X^\top X\alpha - \tilde\vecr}_2^2\\
\end{eqnarray*}
where $X = \br{\Phi_K(\vecz_1),\ldots,\Phi_K(\vecz_k)}$, $\vecr = \br{r_1,\ldots,r_k}^\top$, $P$ is the $k \times k$ diagonal matrix with $P_{ii} = \mu_i$, $\tilde X = X\sqrt{P}$ and $\tilde\vecr = \sqrt{P}\vecr$. The last step follows by the Representer Theorem which tells us that at the optima, $\vecW' = X\alpha$ for some $\alpha \in \R^k$.

Some simple linear algebra shows us that the minimizer $\alpha$ has the form
\begin{eqnarray*}
\alpha &=& \br{X^\top \tilde X \tilde X^\top X + \frac{1}{2mC}X^\top X}^{-1}X^\top \tilde X \tilde\vecr\\
			 &=& \br{GPG + \frac{G}{2mC}}^{-1}GP\vecr\\
			 &=& \br{PG + \frac{I}{2mC}}^{-1}G^{-1}GP\vecr\\
			 &=& \br{PG + \frac{I}{2mC}}^{-1}P\vecr
\end{eqnarray*}
where $G = X^\top X$ is the Gram matrix given by the kernel $K$. In the third step we have assumed that $G$ does not have vanishing eigenvalues which can always be ensured by adding a small positive constant to the diagonal. Thus we have
\[
\br{PG + \frac{I}{2mC}}\alpha = P\vecr
\]
looking at the $i^{th}$ element of both sides we have
\[
\mu_i\sum\limits_{j=1}^k\alpha_jK(\vecz_i,\vecz_j) + \frac{\alpha_i}{m2C} = \mu_ir_i
\]
which gives us $\alpha_i = 2mC\mu_i\br{r_i - \ip{\vecW'}{\Phi_K(\vecz_i)}}$. Now assume, without loss of generality, that the relevance scores are normalized, i.e. $r_i \leq 1$ for all $i$. Thus we have
\[
\frac{1}{2}\norm{\vecW'}_{\H_K}^2 + mC\norm{\tilde X^\top\vecW' - \tilde\vecr}_2^2 \leq \frac{1}{2}\norm{\veczero}_{\H_K}^2 + mC\norm{\tilde X^\top\veczero - \tilde\vecr}_2^2
\]
which gives us $\frac{1}{2}\norm{\vecW'}_{\H_K}^2 \leq mC\norm{\tilde\vecr}_2^2 \leq mC\sum\limits_{i=1}^k \mu_i = mC$ which gives us $\norm{\vecW'} \leq \sqrt{2mC}$. Since the kernel is already a normalized kernel, $\norm{\Phi_K(\vecz_i)} \leq 1$ which gives us, by an application of Cauchy-Schwartz, $\abs{\alpha_i} \leq 2mC\mu_i(1+\sqrt{m2C}) \leq 5\mu_imC\sqrt{mC}$.

If we now establish a weight function over the domain $w_i = \frac{\alpha_i}{\mu_i}$, then $\abs{w_i} \leq 5mC\sqrt{mC}$ and we can show that for all $\vecz$, we have $\ip{\vecW'}{\Phi_K(\vecz)} = \EE{\vecz' \sim \mu}{w(\vecz)K(\vecz,\vecz')}$. Setting $C = \frac{1}{2\epsilon_1\gamma^2}$ finishes the proof.
\end{proof}


\section{Supplementary Experimental Results}
\label{app:exps}
Below we present additional experimental results for regression and ordinal regression problems.

\subsection{Regression Experiments}
We present results on various benchmark datasets considered in Section~\ref{sec:exps} for Gaussian $K(\vecxy) = \exp\br{-\frac{\norm{\vecx-\vecy}_2^2}{2\sigma^2}}$ and Euclidean: $K(\vecxy) = -\norm{\vecx-\vecy}_2^2$ kernels. Following standard practice, we fixed $\sigma$ to be the average pairwise distance between data points in the training set.
\begin{figure*}[t]
	\vspace*{-2ex}
	\centering
	\subfloat[Mean squared error for landmarking (RegLand), sparse landmarking (RegLand-Sp) and kernel regression (KR) for the Gaussian kernel]{
		\label{fig:reg_gaussian}
		\includegraphics[width=0.35\textwidth, angle=0]{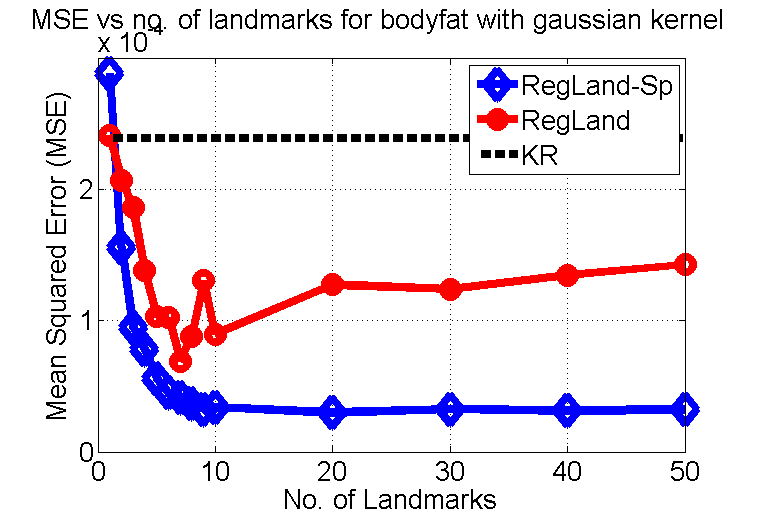}
		\hspace*{-2ex}
		\includegraphics[width=0.35\textwidth, angle=0]{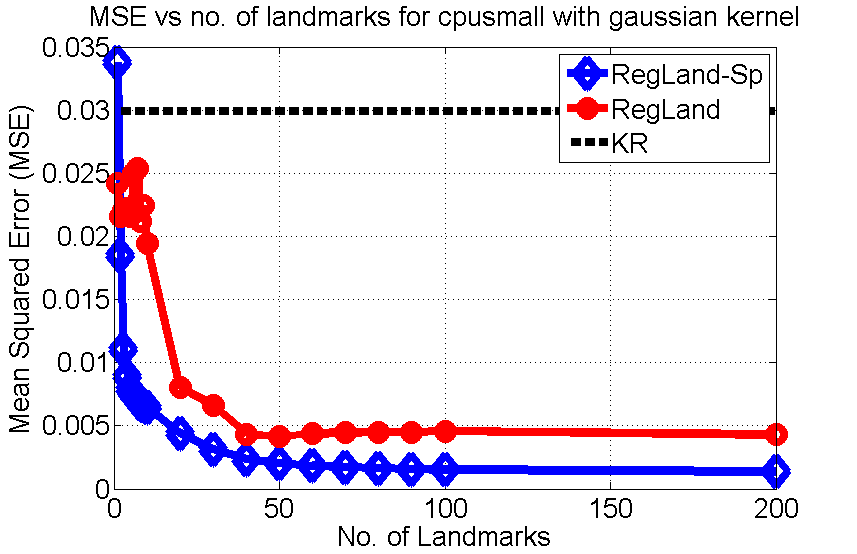}
		\hspace*{-2ex}
		\includegraphics[width=0.35\textwidth, angle=0]{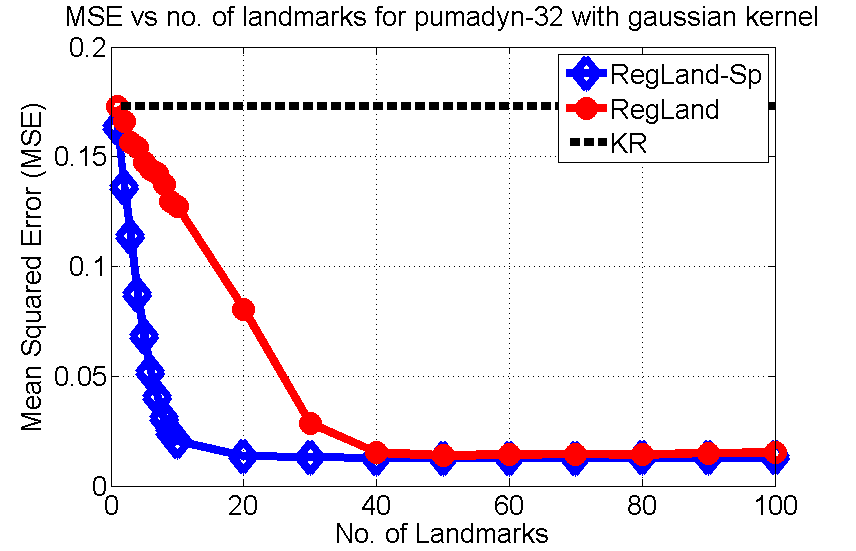}
	}\\\vspace*{-2ex}
	\subfloat[Mean squared error for landmarking (RegLand), sparse landmarking (RegLand-Sp) and kernel regression (KR) for the Euclidean kernel]{
		\label{fig:reg_euclid}
		\includegraphics[width=0.35\textwidth, angle=0]{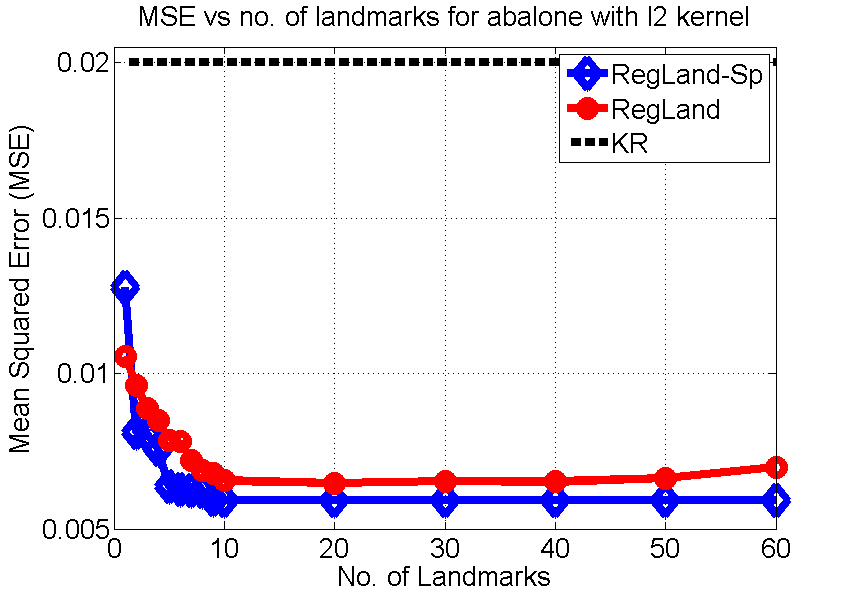}
		\hspace*{-2ex}
		\includegraphics[width=0.35\textwidth, angle=0]{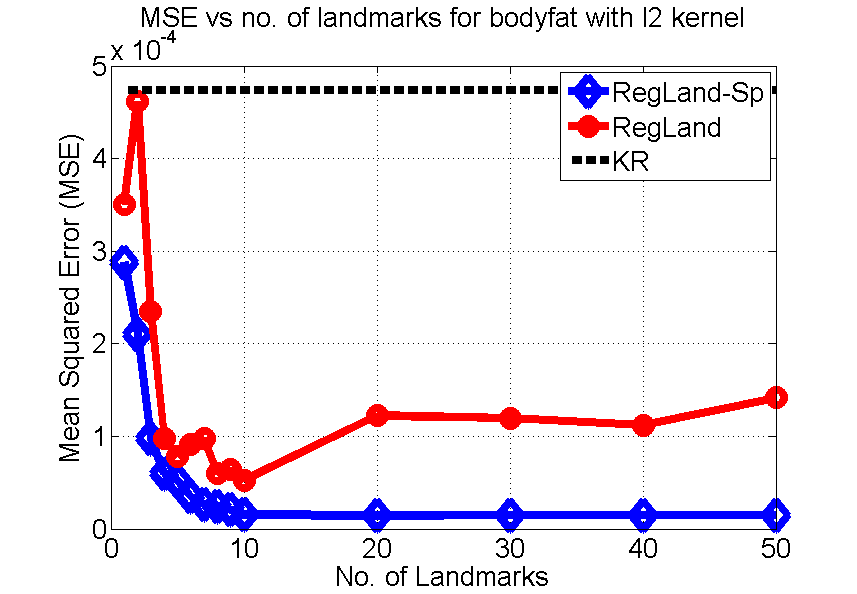}
		\hspace*{-2ex}
		\includegraphics[width=0.35\textwidth, angle=0]{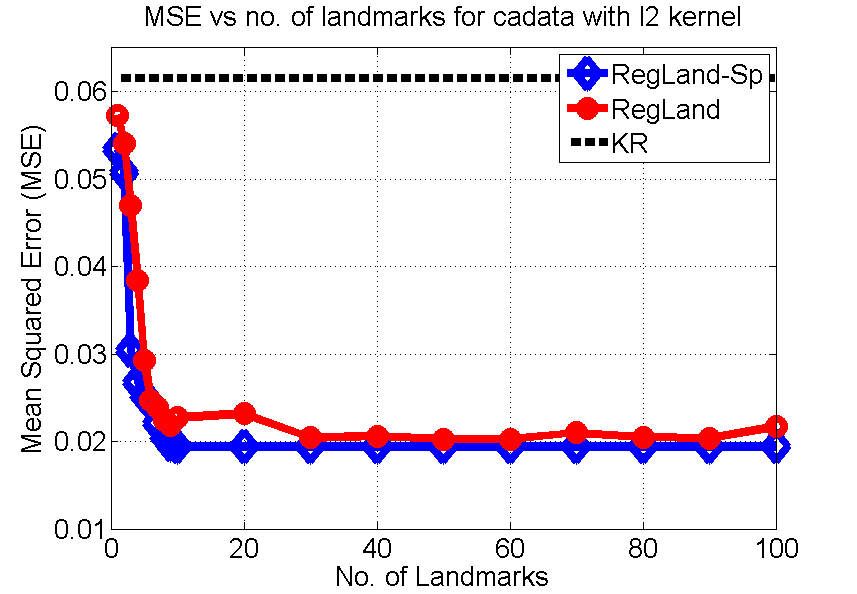}
	}\\\vspace*{-2ex}
	\subfloat[Avg. absolute error for landmarking (ORLand) and kernel regression (KR) on ordinal regression datasets for the Manhattan kernel]{
		\label{fig:ordreg_l1}
		\includegraphics[width=0.35\textwidth, angle=0]{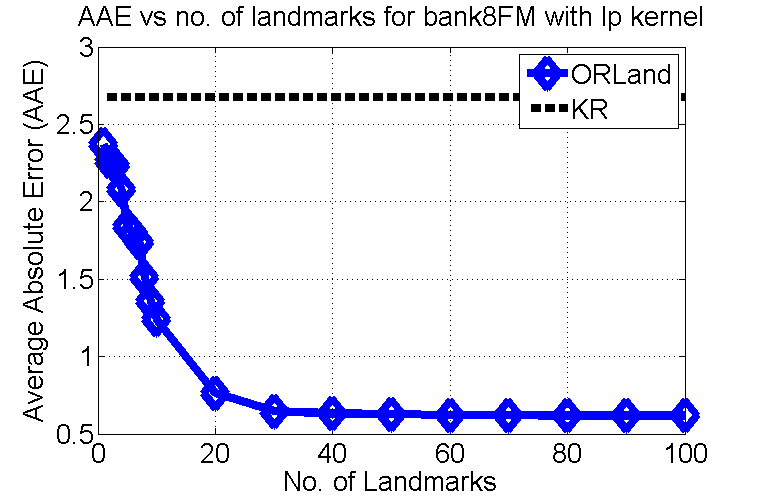}
		\hspace*{-2ex}
		\includegraphics[width=0.35\textwidth, angle=0]{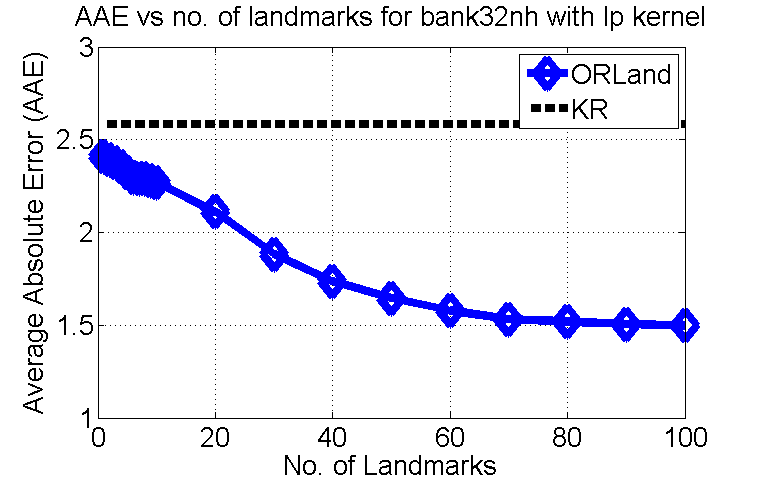}
		\hspace*{-2ex}
		\includegraphics[width=0.35\textwidth, angle=0]{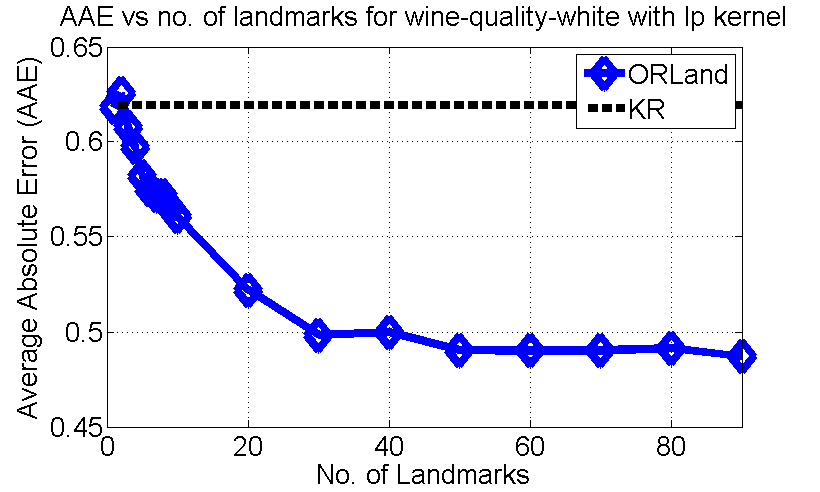}
	}\\\vspace*{-2ex}
	\subfloat[Avg. absolute error for landmarking (ORLand) and kernel regression (KR) on ordinal regression datasets for the Gaussian kernel]{
		\label{fig:ordreg_gaussian}
		\includegraphics[width=0.35\textwidth, angle=0]{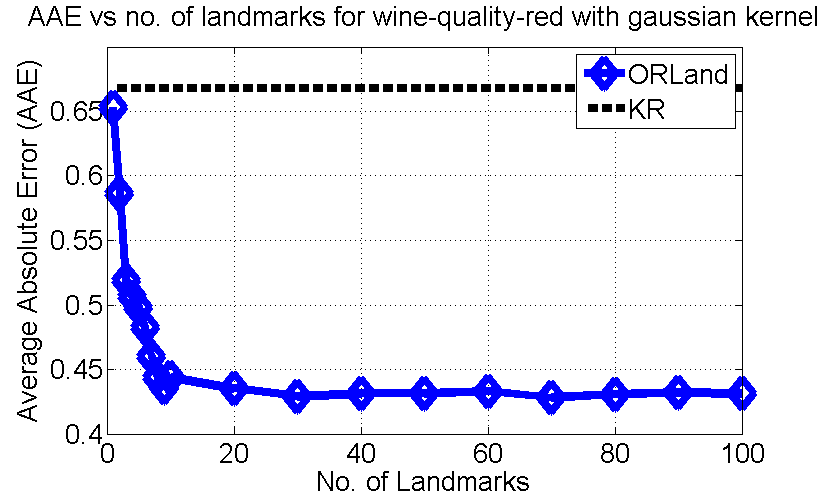}
		\hspace*{-2ex}
		\includegraphics[width=0.35\textwidth, angle=0]{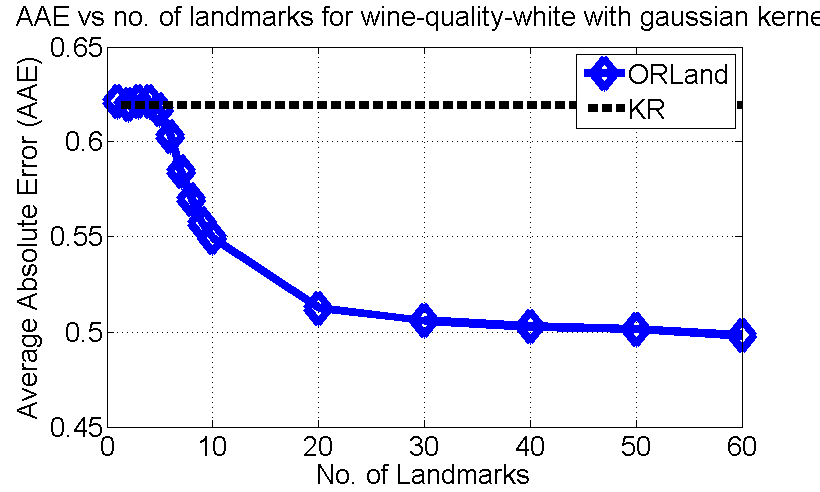}
		\hspace*{-2ex}
		\includegraphics[width=0.35\textwidth, angle=0]{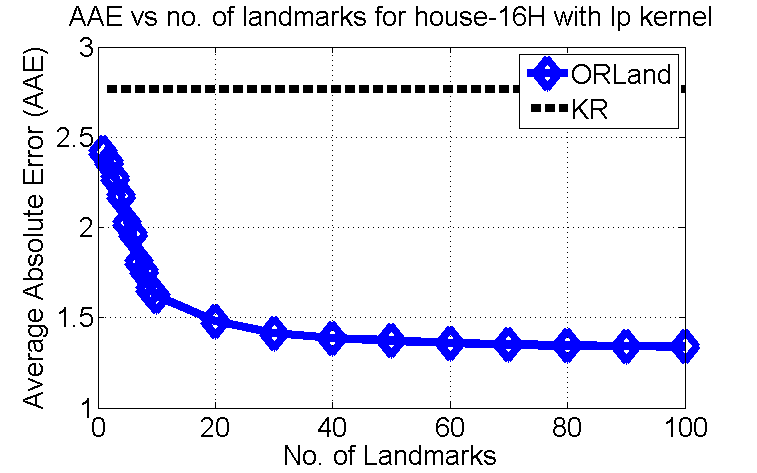}
	}
	\caption{Performance of landmarking algorithms with increasing number of landmarks on real regression (Figures~\ref{fig:reg_gaussian} and \ref{fig:reg_euclid}) and ordinal regression datasets (Figures~\ref{fig:ordreg_l1} and \ref{fig:ordreg_gaussian}) for various kernels.}
	\label{fig:all_graphs_2}
\vspace*{-15pt}
\end{figure*}

\subsection{Ordinal Regression Experiments}
We present results on various benchmark datasets considered in Section~\ref{sec:exps} for Gaussian $K(\vecxy) = \exp\br{-\frac{\norm{\vecx-\vecy}_2^2}{2\sigma^2}}$ and Manhattan: $K(\vecxy) = -\norm{\vecx-\vecy}_1$ kernels.
\end{document}